\documentclass{article}

\usepackage{microtype}
\usepackage{graphicx}
\usepackage{subcaption}
\usepackage{booktabs} 

\usepackage{hyperref}

\usepackage[preprint]{icml2026}

\usepackage{amsmath}
\usepackage{amssymb}
\usepackage{mathtools}
\usepackage{amsthm}

\usepackage{multirow}
\usepackage{wrapfig}

\usepackage[capitalize,noabbrev]{cleveref}

\usepackage[most,skins,theorems]{tcolorbox}

\tcbset{
  aibox/.style={
    width=\linewidth,
    top=8pt,
    bottom=4pt,
    colback=blue!6!white,
    colframe=black,
    colbacktitle=black,
    enhanced,
    center,
    attach boxed title to top left={yshift=-0.1in,xshift=0.15in},
    boxed title style={boxrule=0pt,colframe=white,},
  }
}
\newtcolorbox{AIbox}[2][]{aibox,title=#2,#1}

\theoremstyle{plain}

\theoremstyle{definition}

\theoremstyle{remark}

\usepackage[textsize=tiny]{todonotes}

\icmltitlerunning{PatternKV: Flattening KV Representation Expands Quantization Headroom}

\begin{document}

\twocolumn[
  \icmltitle{PatternKV: Flattening KV Representation Expands Quantization Headroom}



  \icmlsetsymbol{equal}{*}

  \begin{icmlauthorlist}
    \icmlauthor{Ji Zhang}{equal,sch}
    \icmlauthor{Yiwei Li}{equal,sch}
    \icmlauthor{Shaoxiong Feng}{comp}
    \icmlauthor{Peiwen Yuan}{sch}
    \icmlauthor{Xinglin Wang}{sch}
    \icmlauthor{Yueqi Zhang}{sch}
    \icmlauthor{Jiayi Shi}{sch}
    \icmlauthor{Chuyi Tan}{sch}
    \icmlauthor{Boyuan Pan}{comp}
    \icmlauthor{Yao Hu}{comp}
    \icmlauthor{Kan Li}{sch}
  \end{icmlauthorlist}

  \icmlaffiliation{comp}{Xiaohongshu Inc}
  \icmlaffiliation{sch}{School of Computer Science, Beijing Institute of Technology}

  \icmlcorrespondingauthor{Kan Li}{likan@bit.edu.cn}
  \icmlcorrespondingauthor{Shaoxiong Feng}{shaoxiongfeng2023@gmail.com}

  \icmlkeywords{Machine Learning, ICML}

  \vskip 0.3in
]



\printAffiliationsAndNotice{\icmlEqualContribution}

\begin{abstract}

KV cache in autoregressive LLMs eliminates redundant recomputation but has emerged as the dominant memory and bandwidth bottleneck during inference, notably with long contexts and test-time scaling.
KV quantization is a key lever for reducing cache cost, but accuracy drops sharply as the native KV distribution lacks flatness and thus maintains a wide quantization range. 
Prior work focuses on isolating outliers, which caps their error but fails to flatten the overall distribution, leaving performance fragile under low-bit settings.
In this work, we show that the K cache maintains a stable, context-evolving structure, while the V cache carries latent semantic regularities, with both contributing to the organization of vectors into shared patterns.
Building on these insights, we propose \textbf{PatternKV}, a pattern-aligned residual quantization scheme. It mines representative pattern vectors online, aligns each KV vector to its nearest pattern, and quantizes only the residual. This reshaping of the KV distribution flattens the quantization target and narrows its range, thereby improving the fidelity of low-bit KV quantization.
Across long-context and test-time scaling settings on multiple backbones, PatternKV delivers consistent 2-bit gains, with a 0.08\% average 4-bit drop relative to FP16, improves test-time scaling accuracy by 10\% on average, and raises throughput by 1.5× while supporting 1.25× larger batches.

\end{abstract}

\section{Introduction}

Large language models (LLMs) have achieved remarkable performance in various tasks~\citep{GPT4,Qwen2.5, Llama3, Mistral}, yet such performance is grounded in autoregressive decoding. This process relies on a key-value (KV) cache to avoid redundant recomputation, but the cache itself has become a dominant memory and bandwidth bottleneck during inference~\citep{vLLM,FlexGen}. 
This challenge is further compounded by two key drivers: (i) \textbf{long contexts}, prevalent in tasks such as retrieval-augmented generation~\citep{RAG} and long-document processing~\citep{Longformer}; and (ii) \textbf{test-time scaling}, arising from both long chain-of-thought reasoning (depth-oriented expansion)~\citep{s1}, and multi-sample inference like self-consistency~\citep{SC} or tree search~\citep{BeamSearch,REBASE} (breadth-oriented expansion). Taken together, these trends highlight the need for efficient yet high-fidelity KV cache compression in practical LLM deployment.

\begin{figure*}[ht]
  \vskip 0.2in
  \begin{center}
    \centerline{\includegraphics[width=2.0\columnwidth]{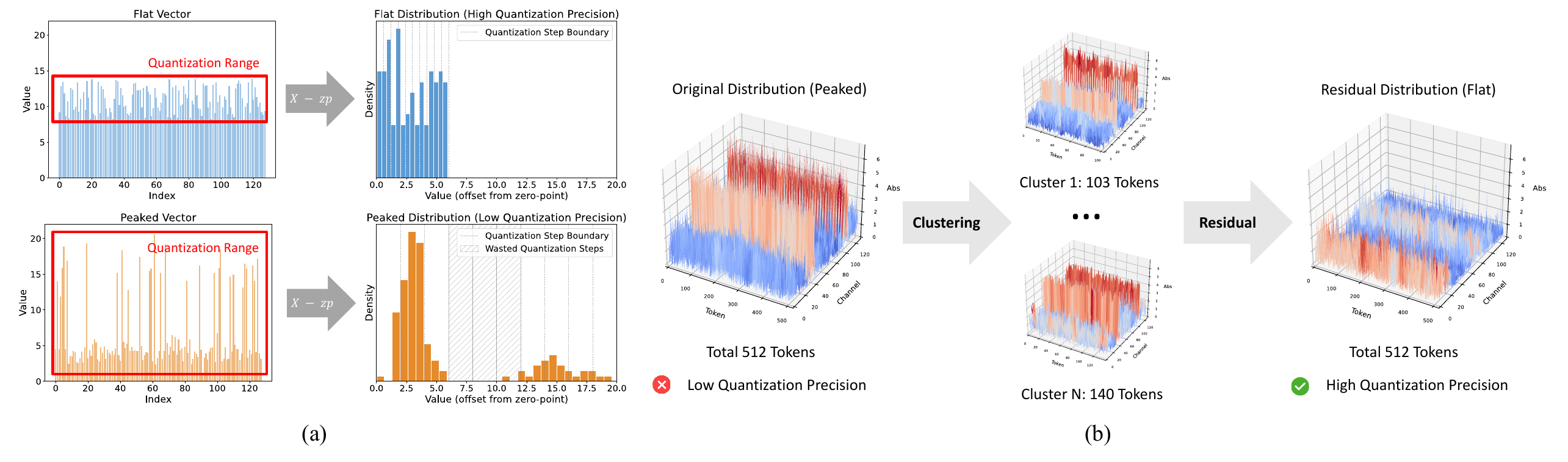}}
    \caption{
      (a) Impact of vector-distribution flatness on KV quantization. Flatter value distributions (top) concentrate samples within a narrower quantization range after zero-point shifting \((X - zp)\), thereby improving the utilization of discrete levels and preserving higher precision at constrained bit widths. In contrast, peaked distributions with outliers (bottom) broaden the required range, leaving many levels underutilized and degrading precision.
      (b) The left figure illustrates the original distribution of the KV vectors, while the right figure depicts the distribution of the residuals obtained after aligning the original vectors with the corresponding pattern vectors. Each pattern vector is the centroid of its cluster.
    }
    \label{fig:var_analysis}
  \end{center}
  
\end{figure*}

Quantization \citep{QuaRot, GPTQ} is a widely adopted approach for KV cache compression, reducing memory footprint via lower-bit KV representations. The effectiveness of KV quantization largely depends on the flatness of the vector distribution (see Figure~\ref{fig:var_analysis}(a) and Section~\ref{sec:var_compose}): flatter distributions yield a narrower quantization range and preserve higher precision under limited bit widths.
In pursuit of this, \citet{KVQuant, GEAR, OTT} handle outliers by storing them with original precision, separated from the main KV distribution to minimize their impact on quantization. Meanwhile, \citet{KIVI} confines outlier-induced quantization error by quantizing key cache per-channel, ensuring the error remains within individual channels. However, these methods are primarily limited to protecting outliers rather than flattening the entire distribution.

In contrast, we tackle the root cause of quantization inefficiency by reshaping the entire KV distribution (Figure~\ref{fig:var_analysis}(b)). 
Guided by a variance–decomposition perspective, we mine common patterns in KV caches, align each vector to its nearest pattern, and quantize only the residuals. This distribution-wide treatment flattens the quantization target, yielding a narrower range and substantially reducing error under low-bit settings.

Specifically, our analysis of KV caches reveals exploitable regularities: the K cache maintains a stable structure that will evolve gradually with context, and the V cache exhibits latent semantic patterns that induce token-consistent structure in the V vectors.
These findings indicate that pattern information can be reliably mined online without calibration corpora or additional tuning. Building on this, we employ clustering to extract representative pattern vectors that capture such common structure. During inference, each KV vector is aligned to its nearest pattern vector and transformed into a residual for quantization, resulting in a markedly flatter distribution. To accommodate the gradual evolution of KV distributions over decoding, we further introduce new pattern vectors on the fly, adaptively tracking shifts and maintaining quantization fidelity.

In summary, our main contributions are as follows:

\begin{itemize}
    \item We introduce a variance–decomposition perspective on KV quantization, which shifts the focus from protecting outliers to flattening the overall distribution.
    
    \item We analyze latent patterns in the K and V caches, revealing stable structural and semantic regularities that motivate pattern-based residualization.

    \item We propose \textbf{PatternKV}, a lightweight, plug-and-play KV quantization scheme that improves low-bit accuracy with minimal overhead.

    \item We evaluate our method against strong baselines across diverse tasks and backbone models.
    In the long-context setting, our approach achieves consistent gains at 2-bit while maintaining \textbf{near-lossless} 4-bit accuracy (only 0.08\% lower than FP16 on average).
    Under test-time scaling, our method achieves a \textbf{10\%} average improvement. In addition, our method achieves a \textbf{1.5×} throughput increase and supports a \textbf{1.25×} larger batch size.
\end{itemize}

\section{Motivations}

\begin{figure*}[ht]
  \vskip 0.2in
  \begin{center}
    \centerline{\includegraphics[width=1.8\columnwidth]{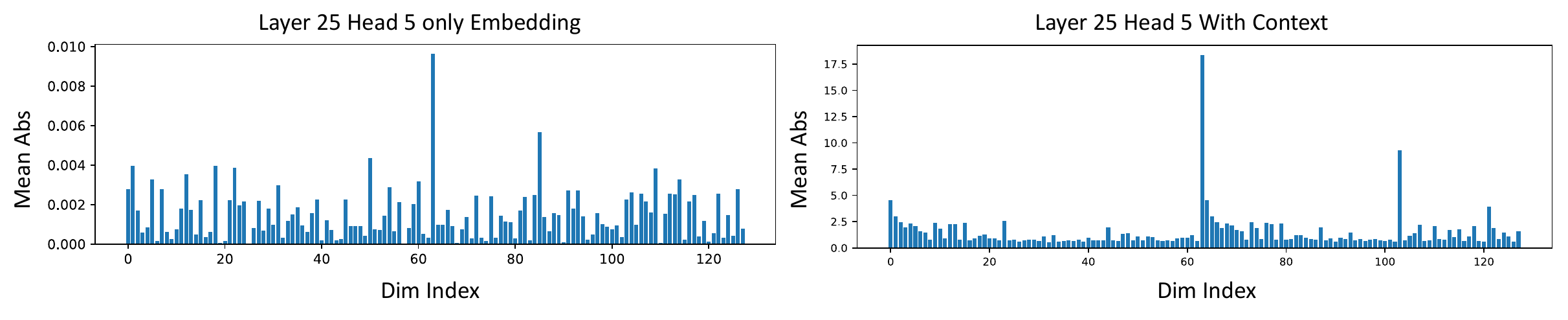}}
    \caption{
      Channel-wise mean absolute value distributions. Left: embedding-only injection; Right: full-input injection. Outlier channels are already evident under embedding-only input, and the full input further enlarges the range and extremes. Additional figures are provided in Appendix~\ref{app:more_figs}.
    }
    \label{fig:k_origin_analysis}
  \end{center}
  \vskip -0.2in
\end{figure*}

\subsection{A Variance Decomposition View of KV Quantization}
\label{sec:var_compose}

In KV cache quantization, asymmetric $n$-bit quantization is typically applied, with each vector $X$ mapped as:
\begin{align}
Q(X) &= \left\lfloor \frac{X - z}{s} \right\rceil, \quad \quad X_{\mathrm{deq}} = s\cdot Q(X) + z,
\end{align}
where $s=\frac{\max(X)-\min(X)}{2^n-1}$ is the scaling factor and $z=\min(X)$ the zero-point, and $\lfloor\cdot\rceil$ denotes rounding to the nearest integer. The scaling factor $s$ critically determines quantization fidelity: a larger $s$ forces more distinct values into the same quantization level, while a smaller $s$ retains finer distinctions. Therefore, flatter KV distributions with smaller ranges $max(X)-min(X)$ yield less distortion under quantization, and we use variance as a natural proxy for this flatness. This leads to the central question: \textbf{how can we reduce the variance of the K and V distributions to improve their quantization fidelity?}

The law of total variance~\citep{var_book} is widely used for analyzing variance reductions~\citep{var_1,var_2}. It states that, given a partition of the data into groups, the total variance can be decomposed into two components: an intra-group term and an inter-group term.
To apply this principle in the KV setting, we can introduce a set of representative pattern vectors $M$ that partition the collection of KV vectors into different clusters. Under this view, the total variance of KV vectors $Z$ decomposes as
\begin{align}
    \operatorname{Var}(Z)
= \underbrace{\mathbb{E}\!\left[\operatorname{Var}(Z \mid M)\right]}_{\text{intra-pattern variance}}
+ \underbrace{\operatorname{Var}\!\left(\mathbb{E}[Z \mid M]\right)}_{\text{inter-pattern variance}}
\end{align}
The second term measures variance across pattern means.
If we fix the pattern set $M$, the inter-pattern term vanishes. So the variance to be quantized reduces to $\mathbb{E}[\operatorname{Var}(Z\mid M)]$. 
Therefore, the key to achieving a flatter quantization target lies in choosing a suitable partition that minimizes intra-pattern variance. In other words, the central challenge shifts from \textbf{reducing error on the raw distribution to selecting pattern vectors $M$ that yield a flatter quantization target.}

\subsection{KV Pattern Analysis}
\label{sec:kv_pattern_analysis}

As established above, selecting a suitable partition is crucial for minimizing variance. We therefore analyze the K and V caches to examine whether they exhibit exploitable latent patterns that can guide the construction of pattern vectors for quantization.
\subsubsection{Origins and Evolution of K Cache Patterns}

Prior work identifies outlier distributions in the K cache~\citep{KIVI,KVQuant}, and we extend this line of evidence with a systematic robustness analysis (Appendix~\ref{app:k_stable_analysis}), which shows that a fixed model’s K cache maintains a stable structure attributable to internal linear mappings and nonlinear activations rather than any particular prompt.
To probe this origin, we run an input–decoupling experiment: for each token, we compare the K cache distribution when propagating only the token embedding to that obtained from the full hidden state carrying context. As shown in Figure~\ref{fig:k_origin_analysis}, outlier channels already appear with embedding-only input, adding context chiefly inflates overall magnitude and dynamic range while leaving the structural pattern intact. The invariance of this pattern across inputs indicates that reliable pattern estimates can be obtained directly from the observed activations, without heavy dependence on corpus-specific calibration. We conclude:

\begin{AIbox}{Insight 1}
    The stable structure in the K cache is primarily model-internal. Context mainly rescales values rather than altering the underlying structure.
\end{AIbox}

\begin{figure*}[ht]
  \vskip 0.2in
  \begin{center}
    \centerline{\includegraphics[width=1.8\columnwidth]{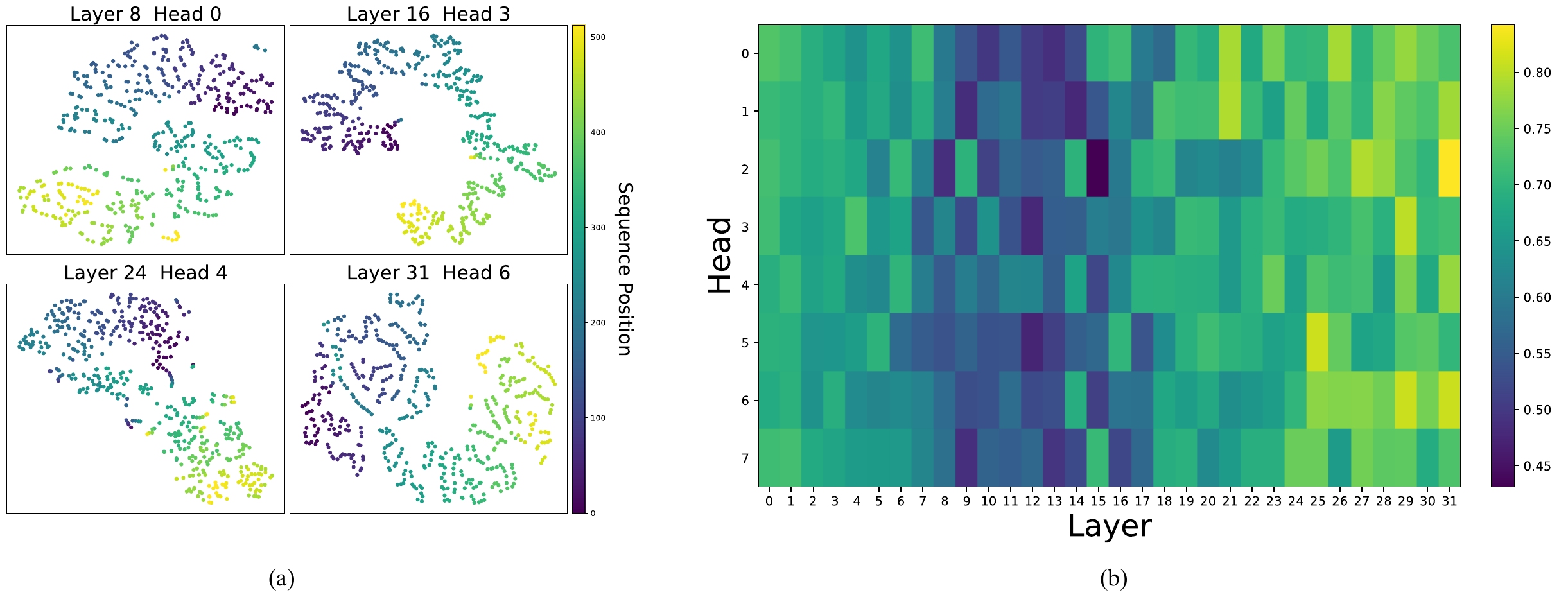}}
    \caption{
      (a) t-SNE visualization of the  of K-cache distributions across attention heads along a single inference trajectory. 
 (b) Illustration of the degree of alignment between V cache clusters and semantic categories. Additional figures are provided in Appendix~\ref{app:more_figs}.
    }
    \label{fig:k_evo_analysis_and_v_analysis}
  \end{center}
  \vskip -0.2in
\end{figure*}
Building on Insight 1, we analyze how the evolving context reshapes the K cache distribution during decoding.
We sample K vectors along a single inference trajectory and visualize them per attention head using t-SNE. As shown in Figure~\ref{fig:k_evo_analysis_and_v_analysis}(a), the K distribution drifts smoothly across decoding steps rather than exhibiting abrupt jumps, and each head follows a distinct trajectory. This behavior is consistent with rotary positional embeddings, which inject relative position after Q and K are formed. Notably, although the marginal distribution evolves, short-range geometry remains stable: nearby tokens along the sequence tend to inhabit similar regions. This local consistency makes it natural to ground pattern estimates in the immediate neighborhood along the trajectory, where local similarity is highest. Hence,

\begin{AIbox}{Insight 2}
    Context and RoPE induce a gradual, head-specific evolution of the K distribution whose direction is difficult to predict.
\end{AIbox}

\subsubsection{Analysis of Latent Pattern in V Cache}
In contrast to the K cache, the V cache shows neither pronounced outliers nor a broad dynamic range, so magnitude-only cues are uninformative. 
Because K does not appear in the output while V does, we instead rely on V’s semantic content to uncover common structure.
We therefore hypothesize a linkage to token semantics. To obtain a conservative estimate of semantic association, we proceed as follows: for each layer and head, we cluster V vectors using KMeans~\citep{kmeans}. For tokens that appear multiple times, we compute their frequency distribution over clusters and define a consistency metric:
\begin{align}
    C_t=\frac{\max_k n_{t,k}}{\sum_k n_{t,k}}
\end{align}
We then aggregate $C_t$ across layers and heads to assess within-cluster cohesion.
As shown in Figure~\ref{fig:k_evo_analysis_and_v_analysis}(b), shallow and deep layers exhibit strong alignment between tokens and cluster assignments, supporting our hypothesis. In the middle layers, the same token spreads across multiple clusters, indicating weaker coupling between V representations and semantics, which hampers the extraction of common structure. Therefore,

\begin{AIbox}{Insight 3}
    The V cache generally exhibits latent semantic patterns, with token--cluster association remaining strong in most layers and attenuating in some middle layers.
\end{AIbox}

\section{Method}

\begin{figure*}[ht]
  \vskip 0.2in
  \begin{center}
    \centerline{\includegraphics[width=1.8\columnwidth]{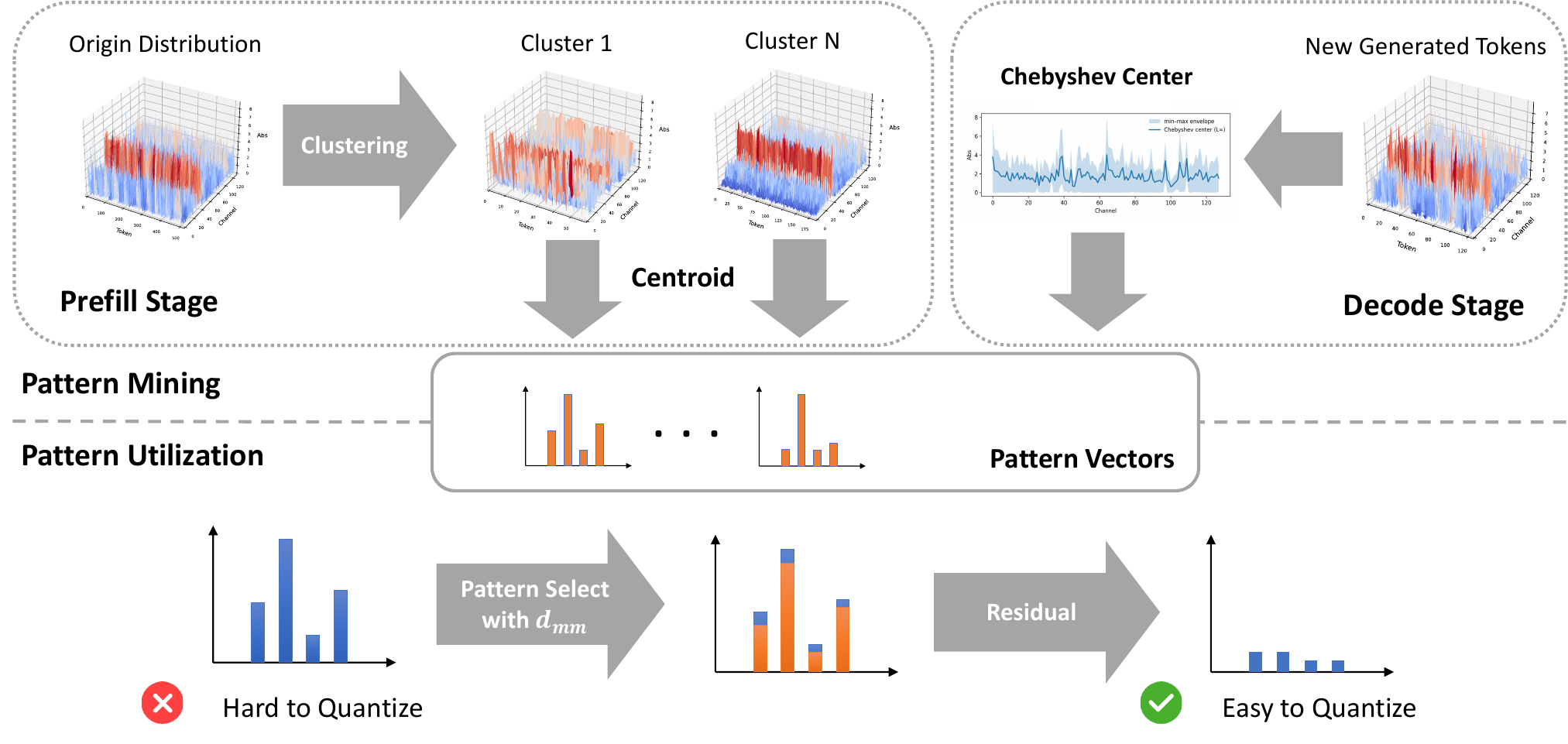}}
    \caption{
      Overview of the PatternKV pipeline: pattern vectors are mined online, KV vectors are aligned to their nearest pattern, and only residuals are quantized.
    }
    \label{fig:method}
  \end{center}
  \vskip -0.2in
\end{figure*}

In light of the previous analysis, we propose \textbf{PatternKV}, a residual quantization pipeline based on pattern alignment, as illustrated in Figure~\ref{fig:method}. 
In the prefill stage, we select pattern vectors online via clustering to minimize within-pattern variance (\textbf{Insight 1}).
In the decode stage, we update the pattern vector to the Chebyshev center to adaptively track the distribution’s gradual evolution (\textbf{Insight 2}) and the potential emergence of new semantic patterns during generation (\textbf{Insight 3}).
For pattern utilization, we assign each KV vector to a pattern using the min–max distance and quantize only the residual, which flattens the target distribution and contracts its dynamic range.
For the V cache, where semantic alignment is weaker in intermediate layers (\textbf{Insight 3}), we further incorporate an adaptive threshold so that flattening provably incurs error no greater than raw quantization. Besides, we provide one theoretical guarantee for the method in Appendix~\ref{app:prove_worst_case}.

\subsection{Pattern Mining}
\label{sec:mining}
\paragraph{Prefilling Stage}

During the prefilling stage, we select and fix a set of pattern vectors so that the variance to be quantized reduces to $\mathbb{E}[\operatorname{Var}(Z\mid M)]$. Our goal is therefore to minimize the within-pattern variance within the chosen partition, with the following optimization objective:
\begin{align}
    \min_{\mathcal{P}_h=P_1,\ldots,P_k}\;\sum_{j=1}^k\;\sum_{\boldsymbol{x_i}\in P_j}\bigl\|\boldsymbol{x_i}-\boldsymbol{\overline{x}_{P_j}}\bigr\|_2^2
\end{align}
Let $P_k$ denote the $k$-th pattern cluster. We optimize the objective using KMeans~\citep{kmeans} under the Euclidean metric and take the centroid of each cluster as its pattern vector. For the $h$-th attention head, the resulting set of pattern vectors is $\mathcal{M}_h=\{M_1,\ldots,M_k\}$. 
Since our objective coincides with the K-means objective, the partition returned at convergence is a local minimizer of the within-pattern variance.

\paragraph{Decoding Stage}
Guided by Insight 2, we update per-head pattern vectors during decoding to adaptively track the distribution’s gradual evolution. Instead of arithmetic means, we use Chebyshev centers computed over each group of KV vectors, which minimize the local quantization range and provide stronger robustness to outliers, thereby aligning better with the asymmetric quantization objective.
Specifically, we use the quantization group window $G_{\text{pattern}}$ to generate new pattern vectors. For the pattern vector $M_h^{new}$ of the $h$-th attention head within this window, we have:
\begin{align}
    M_{h,d}^{new}=\frac{1}{2}\left(\underset{i}{min}\,X_{h,i,d}+\underset{i}{max}\,X_{h,i,d}\right)
\end{align}
Here, $d$ indexes the dimension of the head, and $i$ indexes the $i$-th KV vector within the full-precision window. Once the $M_h^*$ is computed, it is merged into the existing pattern vertor set $\mathcal{M}_h$ for subsequent pattern matching and flattening.

\subsection{Pattern Utilization}
\label{sec:use}
The objective of KV flattening is to minimize the quantization range. To achieve this, we replace direct quantization of raw vectors with residual quantization: each vector is first aligned to a pattern vector, and then only the residual is quantized, which yields a much flatter distribution.

Specifically, we adopt the min–max distance for pattern selection, defined for a vector $\boldsymbol{x}$ and a candidate pattern $\boldsymbol{m}$ as $d_{mm}(\boldsymbol{x},\boldsymbol{m})=\max_i(x_i-m_i)-\min_j(x_j-m_j)$.

During inference, for each KV vector we retrieve its nearest pattern under the $d_{\mathrm{mm}}$ metric. Concretely, the quantized target is the residual aligned to the nearest pattern:
\begin{align}
    M^\star=\underset{M\in\mathcal{M}_h}{argmin}\; d_{\mathrm{mm}}(X,M),\qquad R=X-M^\star
\end{align}
We record the index $k^\star$ together with the quantization parameters. During dequantization, we use this index to retrieve the corresponding pattern vector $M^\star$ and reconstruct the original KV representation by inverting the residualization step.

\subsection{Flattening-Sensitive Adaptive Threshold for V Pattern Utilization}
\label{sec:threshold_for_v}

\begin{table*}[h]
\caption{Overall LongBench results at 2-bit precision. The best and second-best in every column are marked in \textbf{bold} and \underline{underline}, respectively. See Appendix~\ref{app:longbench_int4} for the 4-bit precision results.}
\label{tab:longbench-results}
\small
\renewcommand{\arraystretch}{1.15} 
\begin{center}
\begin{tabular}{p{3cm}p{1.2cm}ccccccc}
\toprule
\multicolumn{1}{c}{\bf Model} &
\multicolumn{1}{c}{\bf Method} &
\multicolumn{1}{c}{\bf MQA} &
\multicolumn{1}{c}{\bf SQA} &
\multicolumn{1}{c}{\bf Summ.} &
\multicolumn{1}{c}{\bf Few-shot} &
\multicolumn{1}{c}{\bf Synth.} &
\multicolumn{1}{c}{\bf Code} &
\multicolumn{1}{c}{\bf Avg} \\
\toprule
\multirow{6}{*}{Llama3.1-8B-Instruct}
& FP16 & 36.63 & 46.56 & 25.54 & 61.16 & 59.99 & 59.42 & 46.59 \\
\cmidrule(lr){2-9}
& KIVI       & \underline{34.86} & \underline{43.96} & 24.98 & \underline{60.35} & 54.43 & 55.53 & 44.33 \\
& ZipCache   & 32.65 & 40.52 & 24.02 & 59.86 & 47.44 & 60.91 & 42.49 \\
& SKVQ       & 34.81 & 42.59 & 24.83 & 59.74 & 52.81 & \underline{61.45} & 44.25 \\
& OTT        & 34.34 & 43.41 & \textbf{25.19} & 59.64 & \underline{55.45} & \textbf{62.48} & \underline{44.84} \\
& PatternKV  & \textbf{35.49} & \textbf{45.08} & \underline{25.12} & \textbf{60.58} & \textbf{57.89} & 56.55 & \textbf{45.33} \\
\midrule
\multirow{6}{*}{Llama3.1-70B-Instruct}
& FP16 & 52.68 & 49.56 & 25.67 & 66.18 & 72.67 & 46.80 & 51.81 \\
\cmidrule(lr){2-9}
& KIVI       & \underline{52.41} & \underline{48.92} & \textbf{25.45} & \underline{65.73} & \underline{72.58} & 46.62 & \underline{51.48} \\
& ZipCache   & 36.98 & 45.44 & 23.28 & 58.57 & 67.92 & \underline{58.37} & 46.55 \\
& SKVQ       & - & - & - & - & - & - & -\\
& OTT        & 40.72 & 47.36 & 24.74 & 60.05 & 68.50 & \textbf{59.97} & 48.43 \\
& PatternKV  & \textbf{52.45} & \textbf{49.19} & \underline{25.21} & \textbf{65.76} & \textbf{72.67} & 47.65 & \textbf{51.61} \\
\midrule
\multirow{3}{*}{Qwen2.5-7B-Instruct}
& FP16 & 38.03 & 45.40 & 23.37 & 59.85 & 58.83 & 62.84 & 46.13 \\
\cmidrule(lr){2-9}
& KIVI       & \underline{35.77} & \underline{42.73} & \textbf{22.80} & \underline{58.13} & \underline{51.50} & \underline{56.25} & \underline{43.08} \\
& PatternKV  & \textbf{36.36} & \textbf{43.93} & \underline{22.77} & \textbf{59.21} & \textbf{55.17} & \textbf{56.67} & \textbf{44.18} \\
\bottomrule
\end{tabular}
\end{center}
\vskip -0.1in
\end{table*}

In mid layers, weak semantic associations can make flattening unreliable. To safeguard against this, we derive an adaptive threshold using a one-sided z-test, deciding whether to utilize the patterns.
Define
\begin{align}
    D=\frac{1}{d}\sum_{i=1}^{d}\bigl(\varepsilon_{\text{raw},i}^{2}-\varepsilon_{\text{flat},i}^{2}\bigr)
\end{align}
where $\varepsilon_{(\cdot),i}$ denotes the error on dimension $i$ and $d$ is the head dimensionality. The null hypothesis is
$$
H_0:\ \mathbb{E}[D]\le 0
$$
, with significance level $\alpha$. Flattening is applied only when $H_0$ is rejected; otherwise, we revert to raw quantization.
Under the high-resolution approximation, the following relationship for the quantization error of the V cache can be derived:
\begin{align}
\mathbb{E}[D]=\frac{\Delta_{raw}^2-\Delta_{flat}^2}{12},
\mathrm{Var}(D)=\frac{\Delta_{raw}^4+\Delta_{flat}^4}{180\,d}
\label{eq:ED_VarD}
\end{align}
Here $\Delta_{(\cdot)}=\frac{R_{(\cdot)}}{2^n-1}$ denotes the n-bit quantization step size. Because the head dimensionality $d$ in modern LLMs is typically large (e.g., 96, 128, 256), by the central limit theorem $D$ is approximately normal. Using the least favorable boundary $\mu=0$, the one-sided $z$-test adopts the rejection region:
\begin{align}
    \frac{\mathbb{E}[D]-0}{\sqrt{\mathrm{Var}(D)}} \ge z_{1-\alpha}
\end{align}
By substituting Eq. \ref{eq:ED_VarD} and the definition of the quantization step size, and defining the contraction ratio as $\rho = R_{\text{flat}}/R_{\text{raw}}$, we obtain the key criterion:
\begin{align}
1-\rho^2\;\ge\;\frac{2\,z_{1-\alpha}}{\sqrt{5\,d}}\;\sqrt{1+\rho^4}
\;\Longleftrightarrow\;
\rho\;\le\;\rho_{*}(d,\alpha)
\end{align}
Here $\rho_*(d,\alpha)$ denotes the solution to the equality in the left-hand criterion. Consequently, it suffices to compute online the quantization ranges before and after flattening, $R_{\text{raw}}$ and $R_{\text{flat}}$, and check whether $\rho=R_{\text{flat}}/R_{\text{raw}}\le \rho_*(d,\alpha)$. If so, we conclude at confidence level $1-\alpha$ that flattening yields a smaller quantization error for the current V vector.

\subsection{System Optimization}

\paragraph{Prefill-stage pattern extraction}
Because each attention head requires its own set of pattern vectors, we implement a fully GPU-parallel KMeans algorithm to minimize the clustering latency during inference.

\paragraph{Decode-stage KV reconstruction}
Reconstructing KV vectors during decoding introduces substantial overhead, so we design two customized CUDA kernels to reduce the cost. The first kernel fuses pattern index restoration for the K cache, dequantization, and the QK matrix multiplication into a single operator. The second kernel is designed to fuse dequantization, pattern index restoration, and attention weight application over V into one operator. Together, these fused kernels significantly reduce decode-stage latency compared with a pure PyTorch implementation.

Following the above system optimizations, we achieve an approximately 15\% increase in peak throughput on a single GPU and a 10\% reduction in per-sample inference latency, further demonstrating the effectiveness of our optimizations.\footnote{Our code is available in our GitHub repository: https://github.com/HCOOOH/PatternKV.}

\section{Experiments}

\begin{table*}[t]
\caption{Overall Results on the Long-CoT Benchmark at 2-bit precision. See Appendix~\ref{app:longcot_int4} for the 4-bit precision results.}
\label{tab:long-cot-results}
\small
\renewcommand{\arraystretch}{1.15} 
\setlength{\tabcolsep}{3pt}
\begin{center}
\begin{tabular}{lllcccccccccc}
\toprule
\multicolumn{1}{c}{\bf Model} &
\multicolumn{1}{c}{\bf Method} &
\multicolumn{2}{c}{\bf AIME 25} &
\multicolumn{2}{c}{\bf AIME 24} &
\multicolumn{2}{c}{\bf AMC 24} &
\multicolumn{2}{c}{\bf AMC 23} \\
\cline{3-10}
&  & \multicolumn{1}{c}{Avg@8} & \multicolumn{1}{c}{Maj@8}
 & \multicolumn{1}{c}{Avg@8} & \multicolumn{1}{c}{Maj@8}
 & \multicolumn{1}{c}{Avg@8} & \multicolumn{1}{c}{Maj@8}
 & \multicolumn{1}{c}{Avg@8} & \multicolumn{1}{c}{Maj@8} \\
\toprule
\multirow{3}{*}{Llama-8B}
  & FP16
  & 32.33 & 37.93
  & 37.93 & 61.55
  & 53.06 & 60.22
  & 85.58 & 90.13 \\
\cmidrule(lr){2-10}
  & KIVI
  & 12.50 & 17.33
  & 10.83 & 14.0
  & 30.52 & \textbf{46.05}
  & 62.19 & 78.0 \\
  & PatternKV
  & \textbf{17.50} & \textbf{27.17}
  & \textbf{16.25} & \textbf{21.33}
  & \textbf{34.44} & 42.11
  & \textbf{63.44} & \textbf{83.13} \\
\midrule

\multirow{3}{*}{Qwen-7B}
  & FP16
  & 38.39 & 52.14
  & 51.67 & 71.67
  & 60.51 & 63.18
  & 90.06 & 94.87 \\
\cmidrule(lr){2-10}
  & KIVI
  & 27.92 & 35.0
  & \textbf{43.75} & \textbf{59.33}
  & 56.11 & 64.0
  & 83.33 & 90.0 \\
  & PatternKV
  & \textbf{30.42} & \textbf{41.33}
  & 42.92 & 53.67
  & \textbf{57.22} & \textbf{65.89}
  & \textbf{84.06} & \textbf{90.26} \\
\midrule

\multirow{3}{*}{Qwen-14B}
  & FP16
  & 45.83 & 63.17
  & 64.58 & 75.83
  & 65.00 & 67.56
  & 92.50 & 95.0 \\
\cmidrule(lr){2-10}
  & KIVI
  & \textbf{37.08} & \textbf{50.5}
  & 45.00 & 60.83
  & 57.22 & 64.67
  & 85.62 & \textbf{92.5} \\
  & PatternKV
  & 35.42 & 46.67
  & \textbf{47.92} & \textbf{68.16}
  & \textbf{62.22} & \textbf{67.78}
  & \textbf{88.12} & \textbf{92.5} \\
\bottomrule

\end{tabular}
\end{center}
\vskip -0.1in
\end{table*}

\subsection{Settings}

\paragraph{Benchmarks}
 
As long contexts and test-time scaling commonly render the KV cache the dominant memory and bandwidth bottleneck during inference, we structure our evaluation into two categories.
For the long-input setting, we use the full LongBench~\citep{LongBench} benchmark, which offers multiple evaluation dimensions with task-specific metrics. LongBench details appear in Appendix \ref{app:longbench}. For reasoning, we consider GSM8K~\citep{GSM8K}, AIME~\citep{AIME2025}, and AMC~\citep{NuminaMath}. GSM8K probes the impact of quantization on chain-of-thought capability, and AIME and AMC evaluate performance under long chain-of-thought scenarios.

\paragraph{Models}
To assess generalization, we evaluate two representative base model families: Llama~\citep{Llama3} and Qwen~\citep{Qwen2.5}. Under the long-CoT setting, we employ Llama and Qwen variants distilled from DeepSeek-R1~\citep{Deepseek-R1} to enable longer chain-of-thought outputs.

\paragraph{Baselines}
Because our method is an online algorithm that requires no offline calibration set, we compare it against online quantization baselines: KIVI~\citep{KIVI}, ZipCache~\citep{ZipCache}, SKVQ~\citep{SKVQ} and OTT~\citep{OTT}.
Detailed experimental settings for the baseline methods are provided in Appendix \ref{app:baseline}.

\paragraph{Quantization Settings}
In all experiments of this section, we fix the number of pattern vectors at $|\mathcal{M}|=32$ and set the quantization group for new pattern selection to $G_{\text{pattern}}=128$. For quantization granularity, we use per-channel quantization for the K cache and per-token quantization for the V cache, matching the KIVI configuration. Since pre-RoPE recomputes rotary positional embeddings at every decoding step, we perform KV pattern selection after RoPE. All experiments were conducted on NVIDIA A100 GPUs with 40 GB of memory.

\subsection{Main Results}

\paragraph{Results on LongBench}

We evaluate on all 21 datasets of LongBench, focusing on two quantization precisions: INT2 and INT4. The 2-bit results in Table~\ref{tab:longbench-results} demonstrate that our approach achieves robust gains over competitive baselines despite the extreme precision constraint. While some baselines achieve notable improvements on code-related tasks yet fail to generalize to other categories, our method provides stable and consistent improvements across task types. Results for INT4 setting are provided in Appendix~\ref{app:longbench_int4}.

\paragraph{Results on Long-CoT Settings}
Test-time scaling improves LLM reasoning through depth-oriented expansion and breadth-oriented expansion, which yields long outputs and substantially increases KV cache usage. To this end, we evaluate models that can generate long chain-of-thought rationales on challenging mathematical benchmarks. For each problem, we generate eight independent responses and report $Avg@8$ (per-sample accuracy averaged over the eight responses) and $Maj@8$ (problem-level accuracy under majority voting across the eight responses). Table~\ref{tab:long-cot-results} reports the INT2 results: prior methods degrade markedly, whereas our method achieves an average 10\% improvement. We also evaluate under the INT4 setting, detailed results are provided in Appendix~\ref{app:longcot_int4}.

\paragraph{Results on GSM8K}

\begin{figure}[ht]
  \vskip 0.2in
  \begin{center}
    \centerline{\includegraphics[width=.9\columnwidth]{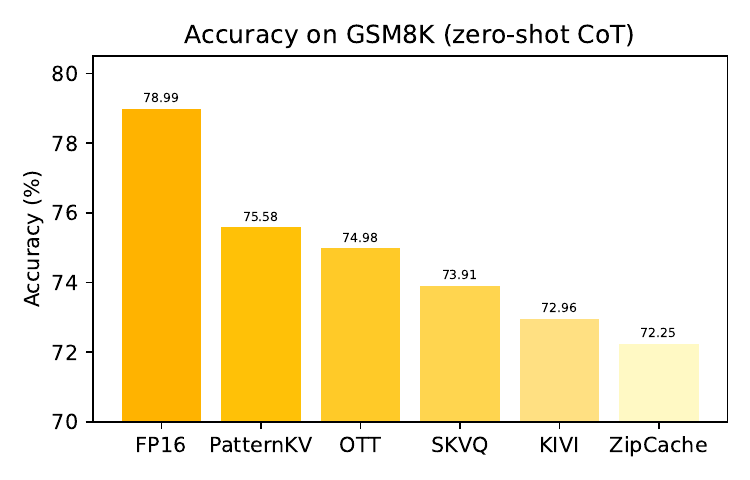}}
    \caption{
      GSM8K accuracy under zero-shot CoT on Llama-3.1-8B-Instruct with an INT2 quantization setting.
    }
    \label{fig:gsm8k_acc}
  \end{center}
  \vskip -0.2in
\end{figure}

We use GSM8K to assess quantization in the non–long-text regime, adopting a zero-shot chain-of-thought paradigm. All results in Fig.~\ref{fig:gsm8k_acc} are obtained under the INT2 quantization setting. Our method reduces accuracy loss in the non–long-text setting. This suggests that preserving the fundamental patterns of KV vectors is critical for maintaining accuracy on reasoning-intensive tasks.

\subsection{Ablation Studies}
We conduct two sets of ablation studies: the first evaluates the contribution of individual components, and the second examines the effect of the number of pattern vectors. Experiments are performed on Llama-3.1-8B-Instruct using LongBench and GSM8K.

\begin{table}[hb]
\centering
\caption{Ablation on Components.}
\label{tab:ablation_components}
\small
\begin{tabular}{lcc}
\toprule
\multicolumn{1}{c}{\bf Component} &
\multicolumn{1}{c}{\bf LongBench Avg} &
\multicolumn{1}{c}{\bf GSM8K} \\
\toprule
KIVI & 44.33 & 72.96 \\
PatternKV & 45.33 & 75.58 \\
\, w/o K Pattern   & 44.53 & 73.91 \\
\, w/o V Pattern   & 44.96 & 74.60 \\
\, w/o New Pattern & 45.37 & 75.49 \\
\, w/o V Threshold & 24.67 & 0.30 \\
\bottomrule
\end{tabular}
\vskip -0.1in
\end{table}

\paragraph{Components}
Table~\ref{tab:ablation_components} shows that each component contributes positively to the overall method. Most notably, removing the adaptive threshold on the V cache leads to substantial performance degradation. This observation corroborates our earlier analysis: because semantic alignment on V varies across layers, a limited number of patterns cannot adequately cover its distribution, the nearest pattern to a given vector may thus be substantially biased, motivating a conservative rejection rule. Despite this, our approach maintains a high level of pattern utilization (about 75\%). For more details, see Appendix~\ref{app:use_rate}. We also find that leveraging patterns on K yields larger gains than on V, consistent with \cite{Quantize_What_Counts}. Under low-bit settings, allocating greater quantization slack to K yields superior quantization benefits.

\begin{table}[ht]
\centering
\caption{Ablation on the number of patterns.}
\label{tab:ablation_pattern_nums}
\small
\begin{tabular}{ccc}
\toprule
\multicolumn{1}{c}{\bf $|\mathcal{M}|$} &
\multicolumn{1}{c}{\bf LongBench Avg} &
\multicolumn{1}{c}{\bf GSM8K} \\
\toprule
KIVI & 44.33 & 72.96 \\
\midrule
2  & 44.57 & 73.72\\
4  & 44.92 & 75.26\\
8  & 44.92 & 75.94\\
16 & 45.28 & 75.20\\
32 & 45.33 & 75.58\\
\bottomrule
\end{tabular}
\vskip -0.1in
\end{table}

\paragraph{The number of patterns}

As shown in Table~\ref{tab:ablation_pattern_nums}, quantization accuracy improves monotonically with the number of patterns. Notably, with $|\mathcal{M}|=4$, we obtain roughly half of the total gains on LongBench and nearly all of the gains on GSM8K. This suggests a task-dependent choice of $|\mathcal{M}|$: long-context tasks benefit from a larger pattern budget to ensure robust coverage, whereas non–long-context tasks achieve comparable accuracy with a smaller number of patterns.

\subsection{Efficiency and Resource Overhead Analysis}

PatternKV is designed to further reduce quantization error, introducing only modest system overhead compared to KIVI.
The primary source of overhead is a one-time pattern-mining step in prefill, and even this single step adds only 6\% to latency.
Despite this, our method delivers \textbf{1.5×} higher throughput than FP16 and increases the maximum single GPU batch size by \textbf{1.25×}. 
While KIVI employs a simpler, highly optimized pipeline which limits its accuracy ceiling, PatternKV delivers significant accuracy improvements with only a minimal reduction in speed.
Moreover, PatternKV consistently outperforms heavier baselines such as ZipCache and SKVQ in both latency and throughput, and remains competitive with OTT while providing clear accuracy improvements. 
For memory overhead, PatternKV is near lossless in practice, increasing peak GPU memory by only about \textbf{0.42\%} compared with KIVI. In contrast, several alternative approaches incur noticeably higher peak-memory pressure, whereas PatternKV keeps the peak GPU memory footprint essentially on par with KIVI while improving generation quality.
A more detailed efficiency analysis is provided in Appendix~\ref{app:efficiency-patternkv}.

\section{Related Work}

KV quantization has developed along three lines of work: (i) outlier-aware compression, where early systems~\citet{FlexGen} showed the feasibility of 4-bit KV but suffered at lower precision; \citet{KIVI} pushed to 2 bits with asymmetric quantization, and later methods~\citet{KVQuant,SKVQ,OTT} mitigated outliers by separating dense and sparse components, constraining error drift, and exempting anomalous tokens; (ii) mixed precision and sensitivity adaptation, where evidence that keys are more fragile than values motivates \citet{AsymKV} to allocate higher precision to K, while \citet{ZipCache} adapts per-token bit widths to capture temporal importance; and (iii) KV sparsification and selective access, where \citet{CQ} compress channels into compact codebooks and \citet{RVQ} stack residual codebooks to approximate KV vectors at a reduced bitrate.
Unlike prior work that partitions or approximates the raw KV distribution, our method explicitly flattens it. By contracting the dynamic range, we unlock greater quantization redundancy and preserve accuracy in low-bit settings.

Complementary to quantization, KV pruning targets redundancy by removing unimportant content before storage. Research follows two lines: (i) sequence-level token selection, where~\citet{StreamingLLM} retain recent tokens via sliding windows, \citet{H2O,Scissorhands} identify heavy hitters using attention scores, and~\citet{PagedEviction,SAGE-KV,TokenSelect,HashEvict,SnapKV} further improve saliency and cache stability via block eviction, one-shot top-k, soft voting, hashing, and representative snapshots; and (ii) structure-level compression, where~\citet{ThinK,KVPruner} prune low-value KV channels and \citet{QUEST} loads only query-relevant KV pages via query-aware metadata. Overall, historical KV caches are highly redundant, with importance concentrated in a small subset of tokens or channels.

\section{Conclusion}

We analyze common patterns in KV caches through a variance–decomposition perspective and introduce PatternKV, a lightweight quantization scheme that reshapes the KV distribution. 
By mining pattern vectors and quantizing residuals, PatternKV reduces intra-pattern variance and contracts the dynamic range, yielding flatter distributions and higher fidelity under low-bit settings.
We establish theoretical support for the method and validate its effectiveness with extensive experiments, while also pointing toward more efficient implementations and system-level integration for broader deployment of LLMs.








\section*{Impact Statement}

This paper presents work whose goal is to advance the field of Machine
Learning. There are many potential societal consequences of our work, none
which we feel must be specifically highlighted here.


\bibliography{Reference}

@article{NuminaMath,
  title={Numinamath: The largest public dataset in ai4maths with 860k pairs of competition math problems and solutions},
  author={Li, Jia and Beeching, Edward and Tunstall, Lewis and Lipkin, Ben and Soletskyi, Roman and Huang, Shengyi and Rasul, Kashif and Yu, Longhui and Jiang, Albert Q and Shen, Ziju and others},
  journal={Hugging Face repository},
  volume={13},
  number={9},
  pages={9},
  year={2024}
}

@article{Longformer,
  author       = {Iz Beltagy and
                  Matthew E. Peters and
                  Arman Cohan},
  title        = {Longformer: The Long-Document Transformer},
  journal      = {CoRR},
  volume       = {abs/2004.05150},
  year         = {2020},
  eprinttype    = {arXiv},
  eprint       = {2004.05150},
  bibsource    = {dblp computer science bibliography, https://dblp.org}
}

@inproceedings{BeamSearch,
  author       = {Yuxi Xie and
                  Kenji Kawaguchi and
                  Yiran Zhao and
                  James Xu Zhao and
                  Min{-}Yen Kan and
                  Junxian He and
                  Michael Qizhe Xie},
  editor       = {Alice Oh and
                  Tristan Naumann and
                  Amir Globerson and
                  Kate Saenko and
                  Moritz Hardt and
                  Sergey Levine},
  title        = {Self-Evaluation Guided Beam Search for Reasoning},
  booktitle    = {Advances in Neural Information Processing Systems 36: Annual Conference
                  on Neural Information Processing Systems 2023, NeurIPS 2023, New Orleans,
                  LA, USA, December 10 - 16, 2023},
  year         = {2023},
  bibsource    = {dblp computer science bibliography, https://dblp.org}
}

@article{s1,
  author       = {Niklas Muennighoff and
                  Zitong Yang and
                  Weijia Shi and
                  Xiang Lisa Li and
                  Li Fei{-}Fei and
                  Hannaneh Hajishirzi and
                  Luke Zettlemoyer and
                  Percy Liang and
                  Emmanuel J. Cand{\`{e}}s and
                  Tatsunori Hashimoto},
  title        = {s1: Simple test-time scaling},
  journal      = {CoRR},
  volume       = {abs/2501.19393},
  year         = {2025},
  doi          = {10.48550/ARXIV.2501.19393},
  eprinttype    = {arXiv},
  eprint       = {2501.19393},
  bibsource    = {dblp computer science bibliography, https://dblp.org}
}

@inproceedings{RAG,
  author       = {Patrick Lewis and
                  Ethan Perez and
                  Aleksandra Piktus and
                  Fabio Petroni and
                  Vladimir Karpukhin and
                  Naman Goyal and
                  Heinrich K{\"{u}}ttler and
                  Mike Lewis and
                  Wen{-}tau Yih and
                  Tim Rockt{\"{a}}schel and
                  Sebastian Riedel and
                  Douwe Kiela},
  editor       = {Hugo Larochelle and
                  Marc'Aurelio Ranzato and
                  Raia Hadsell and
                  Maria{-}Florina Balcan and
                  Hsuan{-}Tien Lin},
  title        = {Retrieval-Augmented Generation for Knowledge-Intensive {NLP} Tasks},
  booktitle    = {Advances in Neural Information Processing Systems 33: Annual Conference
                  on Neural Information Processing Systems 2020, NeurIPS 2020, December
                  6-12, 2020, virtual},
  year         = {2020},
  bibsource    = {dblp computer science bibliography, https://dblp.org}
}

@article{Deepseek-R1,
  author       = {DeepSeek{-}AI and
                  Daya Guo and
                  Dejian Yang and
                  Haowei Zhang and
                  Junxiao Song and
                  Ruoyu Zhang and
                  Runxin Xu and
                  Qihao Zhu and
                  Shirong Ma and
                  Peiyi Wang and
                  Xiao Bi and
                  Xiaokang Zhang and
                  Xingkai Yu and
                  Yu Wu and
                  Z. F. Wu and
                  Zhibin Gou and
                  Zhihong Shao and
                  Zhuoshu Li and
                  Ziyi Gao and
                  Aixin Liu and
                  Bing Xue and
                  Bingxuan Wang and
                  Bochao Wu and
                  Bei Feng and
                  Chengda Lu and
                  Chenggang Zhao and
                  Chengqi Deng and
                  Chenyu Zhang and
                  Chong Ruan and
                  Damai Dai and
                  Deli Chen and
                  Dongjie Ji and
                  Erhang Li and
                  Fangyun Lin and
                  Fucong Dai and
                  Fuli Luo and
                  Guangbo Hao and
                  Guanting Chen and
                  Guowei Li and
                  H. Zhang and
                  Han Bao and
                  Hanwei Xu and
                  Haocheng Wang and
                  Honghui Ding and
                  Huajian Xin and
                  Huazuo Gao and
                  Hui Qu and
                  Hui Li and
                  Jianzhong Guo and
                  Jiashi Li and
                  Jiawei Wang and
                  Jingchang Chen and
                  Jingyang Yuan and
                  Junjie Qiu and
                  Junlong Li and
                  J. L. Cai and
                  Jiaqi Ni and
                  Jian Liang and
                  Jin Chen and
                  Kai Dong and
                  Kai Hu and
                  Kaige Gao and
                  Kang Guan and
                  Kexin Huang and
                  Kuai Yu and
                  Lean Wang and
                  Lecong Zhang and
                  Liang Zhao and
                  Litong Wang and
                  Liyue Zhang and
                  Lei Xu and
                  Leyi Xia and
                  Mingchuan Zhang and
                  Minghua Zhang and
                  Minghui Tang and
                  Meng Li and
                  Miaojun Wang and
                  Mingming Li and
                  Ning Tian and
                  Panpan Huang and
                  Peng Zhang and
                  Qiancheng Wang and
                  Qinyu Chen and
                  Qiushi Du and
                  Ruiqi Ge and
                  Ruisong Zhang and
                  Ruizhe Pan and
                  Runji Wang and
                  R. J. Chen and
                  R. L. Jin and
                  Ruyi Chen and
                  Shanghao Lu and
                  Shangyan Zhou and
                  Shanhuang Chen and
                  Shengfeng Ye and
                  Shiyu Wang and
                  Shuiping Yu and
                  Shunfeng Zhou and
                  Shuting Pan and
                  S. S. Li},
  title        = {DeepSeek-R1: Incentivizing Reasoning Capability in LLMs via Reinforcement
                  Learning},
  journal      = {CoRR},
  volume       = {abs/2501.12948},
  year         = {2025},
  doi          = {10.48550/ARXIV.2501.12948},
  eprinttype    = {arXiv},
  eprint       = {2501.12948},
  bibsource    = {dblp computer science bibliography, https://dblp.org}
}

@misc{AIME2025,
  title = {MathArena: Evaluating LLMs on Uncontaminated Math Competitions},
  author = {Mislav Balunović and Jasper Dekoninck and Ivo Petrov and Nikola Jovanović and Martin Vechev},
  copyright = {MIT},
  publisher = {SRI Lab, ETH Zurich},
  month = feb,
  year = {2025},
}

@article{GSM8K,
  author       = {Karl Cobbe and
                  Vineet Kosaraju and
                  Mohammad Bavarian and
                  Mark Chen and
                  Heewoo Jun and
                  Lukasz Kaiser and
                  Matthias Plappert and
                  Jerry Tworek and
                  Jacob Hilton and
                  Reiichiro Nakano and
                  Christopher Hesse and
                  John Schulman},
  title        = {Training Verifiers to Solve Math Word Problems},
  journal      = {CoRR},
  volume       = {abs/2110.14168},
  year         = {2021},
  eprinttype    = {arXiv},
  eprint       = {2110.14168},
  bibsource    = {dblp computer science bibliography, https://dblp.org}
}

@inproceedings{LongBench,
    title = "{L}ong{B}ench: A Bilingual, Multitask Benchmark for Long Context Understanding",
    author = "Bai, Yushi and Lv, Xin  and Zhang, Jiajie  and Lyu, Hongchang  and
      Tang, Jiankai  and Huang, Zhidian  and Du, Zhengxiao  and Liu, Xiao  and Zeng, Aohan  and Hou, Lei  and Dong, Yuxiao  and Tang, Jie  and Li, Juanzi",
    booktitle = "Proceedings of the 62nd Annual Meeting of the Association for Computational Linguistics (Volume 1: Long Papers)",
    month = aug,
    year = "2024",
    address = "Bangkok, Thailand",
    publisher = "Association for Computational Linguistics",
    doi = "10.18653/v1/2024.acl-long.172",
    pages = "3119--3137",
}

@inproceedings{kmeans,
  title={Some methods of classification and analysis of multivariate observations},
  author={McQueen, James B},
  booktitle={Proc. of 5th Berkeley Symposium on Math. Stat. and Prob.},
  pages={281--297},
  year={1967}
}

@book{var_book,
  title={Introduction to probability},
  author={Blitzstein, Joseph K and Hwang, Jessica},
  year={2019},
  publisher={Chapman and Hall/CRC}
}

@inproceedings{var_1,
  author       = {Stefan Depeweg and
                  Jos{\'{e}} Miguel Hern{\'{a}}ndez{-}Lobato and
                  Finale Doshi{-}Velez and
                  Steffen Udluft},
  editor       = {Jennifer G. Dy and
                  Andreas Krause},
  title        = {Decomposition of Uncertainty in Bayesian Deep Learning for Efficient
                  and Risk-sensitive Learning},
  booktitle    = {Proceedings of the 35th International Conference on Machine Learning,
                  {ICML} 2018, Stockholmsm{\"{a}}ssan, Stockholm, Sweden, July
                  10-15, 2018},
  series       = {Proceedings of Machine Learning Research},
  volume       = {80},
  pages        = {1192--1201},
  publisher    = {{PMLR}},
  year         = {2018},
  bibsource    = {dblp computer science bibliography, https://dblp.org}
}

@inproceedings{var_2,
  author       = {Balaji Lakshminarayanan and
                  Alexander Pritzel and
                  Charles Blundell},
  editor       = {Isabelle Guyon and
                  Ulrike von Luxburg and
                  Samy Bengio and
                  Hanna M. Wallach and
                  Rob Fergus and
                  S. V. N. Vishwanathan and
                  Roman Garnett},
  title        = {Simple and Scalable Predictive Uncertainty Estimation using Deep Ensembles},
  booktitle    = {Advances in Neural Information Processing Systems 30: Annual Conference
                  on Neural Information Processing Systems 2017, December 4-9, 2017,
                  Long Beach, CA, {USA}},
  pages        = {6402--6413},
  year         = {2017},
  bibsource    = {dblp computer science bibliography, https://dblp.org}
}

@article{GPTQ,
  author       = {Elias Frantar and
                  Saleh Ashkboos and
                  Torsten Hoefler and
                  Dan Alistarh},
  title        = {{GPTQ:} Accurate Post-Training Quantization for Generative Pre-trained
                  Transformers},
  journal      = {CoRR},
  volume       = {abs/2210.17323},
  year         = {2022},
  doi          = {10.48550/ARXIV.2210.17323},
  eprinttype    = {arXiv},
  eprint       = {2210.17323},
  bibsource    = {dblp computer science bibliography, https://dblp.org}
}

@inproceedings{REBASE,
  author       = {Yangzhen Wu and
                  Zhiqing Sun and
                  Shanda Li and
                  Sean Welleck and
                  Yiming Yang},
  title        = {Inference Scaling Laws: An Empirical Analysis of Compute-Optimal Inference
                  for {LLM} Problem-Solving},
  booktitle    = {The Thirteenth International Conference on Learning Representations,
                  {ICLR} 2025, Singapore, April 24-28, 2025},
  publisher    = {OpenReview.net},
  year         = {2025},
  bibsource    = {dblp computer science bibliography, https://dblp.org}
}

@inproceedings{SC,
  author       = {Xuezhi Wang and
                  Jason Wei and
                  Dale Schuurmans and
                  Quoc V. Le and
                  Ed H. Chi and
                  Sharan Narang and
                  Aakanksha Chowdhery and
                  Denny Zhou},
  title        = {Self-Consistency Improves Chain of Thought Reasoning in Language Models},
  booktitle    = {The Eleventh International Conference on Learning Representations,
                  {ICLR} 2023, Kigali, Rwanda, May 1-5, 2023},
  publisher    = {OpenReview.net},
  year         = {2023},
  bibsource    = {dblp computer science bibliography, https://dblp.org}
}

@inproceedings{vLLM,
  author       = {Woosuk Kwon and
                  Zhuohan Li and
                  Siyuan Zhuang and
                  Ying Sheng and
                  Lianmin Zheng and
                  Cody Hao Yu and
                  Joseph Gonzalez and
                  Hao Zhang and
                  Ion Stoica},
  editor       = {Jason Flinn and
                  Margo I. Seltzer and
                  Peter Druschel and
                  Antoine Kaufmann and
                  Jonathan Mace},
  title        = {Efficient Memory Management for Large Language Model Serving with
                  PagedAttention},
  booktitle    = {Proceedings of the 29th Symposium on Operating Systems Principles,
                  {SOSP} 2023, Koblenz, Germany, October 23-26, 2023},
  pages        = {611--626},
  publisher    = {{ACM}},
  year         = {2023},
  doi          = {10.1145/3600006.3613165},
  bibsource    = {dblp computer science bibliography, https://dblp.org}
}

@article{GPT4,
  author       = {OpenAI},
  title        = {{GPT-4} Technical Report},
  journal      = {CoRR},
  volume       = {abs/2303.08774},
  year         = {2023},
  doi          = {10.48550/ARXIV.2303.08774},
  eprinttype    = {arXiv},
  eprint       = {2303.08774},
  bibsource    = {dblp computer science bibliography, https://dblp.org}
}

@article{Llama3,
  author       = {Abhimanyu Dubey and
                  Abhinav Jauhri and
                  Abhinav Pandey and
                  Abhishek Kadian and
                  Ahmad Al{-}Dahle and
                  Aiesha Letman and
                  Akhil Mathur and
                  Alan Schelten and
                  Amy Yang and
                  Angela Fan and
                  Anirudh Goyal and
                  Anthony Hartshorn and
                  Aobo Yang and
                  Archi Mitra and
                  Archie Sravankumar and
                  Artem Korenev and
                  Arthur Hinsvark and
                  Arun Rao and
                  Aston Zhang and
                  Aur{\'{e}}lien Rodriguez and
                  Austen Gregerson and
                  Ava Spataru and
                  Baptiste Rozi{\`{e}}re and
                  Bethany Biron and
                  Binh Tang and
                  Bobbie Chern and
                  Charlotte Caucheteux and
                  Chaya Nayak and
                  Chloe Bi and
                  Chris Marra and
                  Chris McConnell and
                  Christian Keller and
                  Christophe Touret and
                  Chunyang Wu and
                  Corinne Wong and
                  Cristian Canton Ferrer and
                  Cyrus Nikolaidis and
                  Damien Allonsius and
                  Daniel Song and
                  Danielle Pintz and
                  Danny Livshits and
                  David Esiobu and
                  Dhruv Choudhary and
                  Dhruv Mahajan and
                  Diego Garcia{-}Olano and
                  Diego Perino and
                  Dieuwke Hupkes and
                  Egor Lakomkin and
                  Ehab AlBadawy and
                  Elina Lobanova and
                  Emily Dinan and
                  Eric Michael Smith and
                  Filip Radenovic and
                  Frank Zhang and
                  Gabriel Synnaeve and
                  Gabrielle Lee and
                  Georgia Lewis Anderson and
                  Graeme Nail and
                  Gr{\'{e}}goire Mialon and
                  Guan Pang and
                  Guillem Cucurell and
                  Hailey Nguyen and
                  Hannah Korevaar and
                  Hu Xu and
                  Hugo Touvron and
                  Iliyan Zarov and
                  Imanol Arrieta Ibarra and
                  Isabel M. Kloumann and
                  Ishan Misra and
                  Ivan Evtimov and
                  Jade Copet and
                  Jaewon Lee and
                  Jan Geffert and
                  Jana Vranes and
                  Jason Park and
                  Jay Mahadeokar and
                  Jeet Shah and
                  Jelmer van der Linde and
                  Jennifer Billock and
                  Jenny Hong and
                  Jenya Lee and
                  Jeremy Fu and
                  Jianfeng Chi and
                  Jianyu Huang and
                  Jiawen Liu and
                  Jie Wang and
                  Jiecao Yu and
                  Joanna Bitton and
                  Joe Spisak and
                  Jongsoo Park and
                  Joseph Rocca and
                  Joshua Johnstun and
                  Joshua Saxe and
                  Junteng Jia and
                  Kalyan Vasuden Alwala and
                  Kartikeya Upasani and
                  Kate Plawiak and
                  Ke Li and
                  Kenneth Heafield and
                  Kevin Stone and
                  et al.},
  title        = {The Llama 3 Herd of Models},
  journal      = {CoRR},
  volume       = {abs/2407.21783},
  year         = {2024},
  doi          = {10.48550/ARXIV.2407.21783},
  eprinttype    = {arXiv},
  eprint       = {2407.21783},
  bibsource    = {dblp computer science bibliography, https://dblp.org}
}

@article{Qwen2.5,
  author       = {An Yang and
                  Baosong Yang and
                  Beichen Zhang and
                  Binyuan Hui and
                  Bo Zheng and
                  Bowen Yu and
                  Chengyuan Li and
                  Dayiheng Liu and
                  Fei Huang and
                  Haoran Wei and
                  Huan Lin and
                  Jian Yang and
                  Jianhong Tu and
                  Jianwei Zhang and
                  Jianxin Yang and
                  Jiaxi Yang and
                  Jingren Zhou and
                  Junyang Lin and
                  Kai Dang and
                  Keming Lu and
                  Keqin Bao and
                  Kexin Yang and
                  Le Yu and
                  Mei Li and
                  Mingfeng Xue and
                  Pei Zhang and
                  Qin Zhu and
                  Rui Men and
                  Runji Lin and
                  Tianhao Li and
                  Tingyu Xia and
                  Xingzhang Ren and
                  Xuancheng Ren and
                  Yang Fan and
                  Yang Su and
                  Yichang Zhang and
                  Yu Wan and
                  Yuqiong Liu and
                  Zeyu Cui and
                  Zhenru Zhang and
                  Zihan Qiu},
  title        = {Qwen2.5 Technical Report},
  journal      = {CoRR},
  volume       = {abs/2412.15115},
  year         = {2024},
  doi          = {10.48550/ARXIV.2412.15115},
  eprinttype    = {arXiv},
  eprint       = {2412.15115},
  bibsource    = {dblp computer science bibliography, https://dblp.org}
}

@article{Mistral,
  author       = {Albert Q. Jiang and
                  Alexandre Sablayrolles and
                  Arthur Mensch and
                  Chris Bamford and
                  Devendra Singh Chaplot and
                  Diego de Las Casas and
                  Florian Bressand and
                  Gianna Lengyel and
                  Guillaume Lample and
                  Lucile Saulnier and
                  L{\'{e}}lio Renard Lavaud and
                  Marie{-}Anne Lachaux and
                  Pierre Stock and
                  Teven Le Scao and
                  Thibaut Lavril and
                  Thomas Wang and
                  Timoth{\'{e}}e Lacroix and
                  William El Sayed},
  title        = {Mistral 7B},
  journal      = {CoRR},
  volume       = {abs/2310.06825},
  year         = {2023},
  doi          = {10.48550/ARXIV.2310.06825},
  eprinttype    = {arXiv},
  eprint       = {2310.06825},
  bibsource    = {dblp computer science bibliography, https://dblp.org}
}

@inproceedings{KIVI,
  author       = {Zirui Liu and
                  Jiayi Yuan and
                  Hongye Jin and
                  Shaochen (Henry) Zhong and
                  Zhaozhuo Xu and
                  Vladimir Braverman and
                  Beidi Chen and
                  Xia Hu},
  title        = {{KIVI:} {A} Tuning-Free Asymmetric 2bit Quantization for {KV} Cache},
  booktitle    = {Forty-first International Conference on Machine Learning, {ICML} 2024,
                  Vienna, Austria, July 21-27, 2024},
  publisher    = {OpenReview.net},
  year         = {2024},
  bibsource    = {dblp computer science bibliography, https://dblp.org}
}

@inproceedings{KVQuant,
  author       = {Coleman Hooper and
                  Sehoon Kim and
                  Hiva Mohammadzadeh and
                  Michael W. Mahoney and
                  Yakun Sophia Shao and
                  Kurt Keutzer and
                  Amir Gholami},
  editor       = {Amir Globersons and
                  Lester Mackey and
                  Danielle Belgrave and
                  Angela Fan and
                  Ulrich Paquet and
                  Jakub M. Tomczak and
                  Cheng Zhang},
  title        = {KVQuant: Towards 10 Million Context Length {LLM} Inference with {KV}
                  Cache Quantization},
  booktitle    = {Advances in Neural Information Processing Systems 38: Annual Conference
                  on Neural Information Processing Systems 2024, NeurIPS 2024, Vancouver,
                  BC, Canada, December 10 - 15, 2024},
  year         = {2024},
  bibsource    = {dblp computer science bibliography, https://dblp.org}
}

@article{GEAR,
  author       = {Hao Kang and
                  Qingru Zhang and
                  Souvik Kundu and
                  Geonhwa Jeong and
                  Zaoxing Liu and
                  Tushar Krishna and
                  Tuo Zhao},
  title        = {{GEAR:} An Efficient {KV} Cache Compression Recipe for Near-Lossless
                  Generative Inference of {LLM}},
  journal      = {CoRR},
  volume       = {abs/2403.05527},
  year         = {2024},
  doi          = {10.48550/ARXIV.2403.05527},
  eprinttype    = {arXiv},
  eprint       = {2403.05527},
  bibsource    = {dblp computer science bibliography, https://dblp.org}
}

@inproceedings{OTT,
  author       = {Yi Su and
                  Yuechi Zhou and
                  Quantong Qiu and
                  Juntao Li and
                  Qingrong Xia and
                  Ping Li and
                  Xinyu Duan and
                  Zhefeng Wang and
                  Min Zhang},
  editor       = {Wanxiang Che and
                  Joyce Nabende and
                  Ekaterina Shutova and
                  Mohammad Taher Pilehvar},
  title        = {Accurate {KV} Cache Quantization with Outlier Tokens Tracing},
  booktitle    = {Proceedings of the 63rd Annual Meeting of the Association for Computational
                  Linguistics (Volume 1: Long Papers), {ACL} 2025, Vienna, Austria,
                  July 27 - August 1, 2025},
  pages        = {12895--12915},
  publisher    = {Association for Computational Linguistics},
  year         = {2025},
  bibsource    = {dblp computer science bibliography, https://dblp.org}
}

@inproceedings{StreamingLLM,
  author       = {Guangxuan Xiao and
                  Yuandong Tian and
                  Beidi Chen and
                  Song Han and
                  Mike Lewis},
  title        = {Efficient Streaming Language Models with Attention Sinks},
  booktitle    = {The Twelfth International Conference on Learning Representations,
                  {ICLR} 2024, Vienna, Austria, May 7-11, 2024},
  publisher    = {OpenReview.net},
  year         = {2024},
  bibsource    = {dblp computer science bibliography, https://dblp.org}
}

@inproceedings{H2O,
  author       = {Zhenyu Zhang and
                  Ying Sheng and
                  Tianyi Zhou and
                  Tianlong Chen and
                  Lianmin Zheng and
                  Ruisi Cai and
                  Zhao Song and
                  Yuandong Tian and
                  Christopher R{\'{e}} and
                  Clark W. Barrett and
                  Zhangyang Wang and
                  Beidi Chen},
  editor       = {Alice Oh and
                  Tristan Naumann and
                  Amir Globerson and
                  Kate Saenko and
                  Moritz Hardt and
                  Sergey Levine},
  title        = {{H2O:} Heavy-Hitter Oracle for Efficient Generative Inference of Large
                  Language Models},
  booktitle    = {Advances in Neural Information Processing Systems 36: Annual Conference
                  on Neural Information Processing Systems 2023, NeurIPS 2023, New Orleans,
                  LA, USA, December 10 - 16, 2023},
  year         = {2023},
  bibsource    = {dblp computer science bibliography, https://dblp.org}
}

@inproceedings{Scissorhands,
  author       = {Zichang Liu and
                  Aditya Desai and
                  Fangshuo Liao and
                  Weitao Wang and
                  Victor Xie and
                  Zhaozhuo Xu and
                  Anastasios Kyrillidis and
                  Anshumali Shrivastava},
  editor       = {Alice Oh and
                  Tristan Naumann and
                  Amir Globerson and
                  Kate Saenko and
                  Moritz Hardt and
                  Sergey Levine},
  title        = {Scissorhands: Exploiting the Persistence of Importance Hypothesis
                  for {LLM} {KV} Cache Compression at Test Time},
  booktitle    = {Advances in Neural Information Processing Systems 36: Annual Conference
                  on Neural Information Processing Systems 2023, NeurIPS 2023, New Orleans,
                  LA, USA, December 10 - 16, 2023},
  year         = {2023},
  bibsource    = {dblp computer science bibliography, https://dblp.org}
}

@article{PagedEviction,
  title={PagedEviction: Structured Block-wise KV Cache Pruning for Efficient Large Language Model Inference},
  author={Chitty-Venkata, Krishna Teja and Ye, Jie and Sun, Xian-He and Kougkas, Anthony and Emani, Murali and Vishwanath, Venkatram and Nicolae, Bogdan},
  journal={arXiv preprint arXiv:2509.04377},
  year={2025}
}

@article{SAGE-KV,
  author       = {Guangtao Wang and
                  Shubhangi Upasani and
                  Chen Wu and
                  Darshan Gandhi and
                  Jonathan Li and
                  Changran Hu and
                  Bo Li and
                  Urmish Thakker},
  title        = {LLMs Know What to Drop: Self-Attention Guided {KV} Cache Eviction
                  for Efficient Long-Context Inference},
  journal      = {CoRR},
  volume       = {abs/2503.08879},
  year         = {2025},
  doi          = {10.48550/ARXIV.2503.08879},
  eprinttype    = {arXiv},
  eprint       = {2503.08879},
  bibsource    = {dblp computer science bibliography, https://dblp.org}
}

@article{TokenSelect,
  author       = {Wei Wu and
                  Zhuoshi Pan and
                  Chao Wang and
                  Liyi Chen and
                  Yunchu Bai and
                  Kun Fu and
                  Zheng Wang and
                  Hui Xiong},
  title        = {TokenSelect: Efficient Long-Context Inference and Length Extrapolation
                  for LLMs via Dynamic Token-Level {KV} Cache Selection},
  journal      = {CoRR},
  volume       = {abs/2411.02886},
  year         = {2024},
  doi          = {10.48550/ARXIV.2411.02886},
  eprinttype    = {arXiv},
  eprint       = {2411.02886},
  bibsource    = {dblp computer science bibliography, https://dblp.org}
}

@article{HashEvict,
  author       = {Minghui Liu and
                  Tahseen Rabbani and
                  Tony O'Halloran and
                  Ananth Sankaralingam and
                  Mary{-}Anne Hartley and
                  Brian J. Gravelle and
                  Furong Huang and
                  Cornelia Ferm{\"{u}}ller and
                  Yiannis Aloimonos},
  title        = {HashEvict: {A} Pre-Attention {KV} Cache Eviction Strategy using Locality-Sensitive
                  Hashing},
  journal      = {CoRR},
  volume       = {abs/2412.16187},
  year         = {2024},
  doi          = {10.48550/ARXIV.2412.16187},
  eprinttype    = {arXiv},
  eprint       = {2412.16187},
  bibsource    = {dblp computer science bibliography, https://dblp.org}
}

@inproceedings{SnapKV,
  author       = {Yuhong Li and
                  Yingbing Huang and
                  Bowen Yang and
                  Bharat Venkitesh and
                  Acyr Locatelli and
                  Hanchen Ye and
                  Tianle Cai and
                  Patrick Lewis and
                  Deming Chen},
  editor       = {Amir Globersons and
                  Lester Mackey and
                  Danielle Belgrave and
                  Angela Fan and
                  Ulrich Paquet and
                  Jakub M. Tomczak and
                  Cheng Zhang},
  title        = {SnapKV: {LLM} Knows What You are Looking for Before Generation},
  booktitle    = {Advances in Neural Information Processing Systems 38: Annual Conference
                  on Neural Information Processing Systems 2024, NeurIPS 2024, Vancouver,
                  BC, Canada, December 10 - 15, 2024},
  year         = {2024},
  bibsource    = {dblp computer science bibliography, https://dblp.org}
}

@inproceedings{ThinK,
  author       = {Yuhui Xu and
                  Zhanming Jie and
                  Hanze Dong and
                  Lei Wang and
                  Xudong Lu and
                  Aojun Zhou and
                  Amrita Saha and
                  Caiming Xiong and
                  Doyen Sahoo},
  title        = {ThinK: Thinner Key Cache by Query-Driven Pruning},
  booktitle    = {The Thirteenth International Conference on Learning Representations,
                  {ICLR} 2025, Singapore, April 24-28, 2025},
  publisher    = {OpenReview.net},
  year         = {2025},
  bibsource    = {dblp computer science bibliography, https://dblp.org}
}

@inproceedings{KVPruner,
  author       = {Bo Lv and
                  Quan Zhou and
                  Xuanang Ding and
                  Yan Wang and
                  Zeming Ma},
  title        = {KVPruner: Structural Pruning for Faster and Memory-Efficient Large
                  Language Models},
  booktitle    = {2025 {IEEE} International Conference on Acoustics, Speech and Signal
                  Processing, {ICASSP} 2025, Hyderabad, India, April 6-11, 2025},
  pages        = {1--5},
  publisher    = {{IEEE}},
  year         = {2025},
  doi          = {10.1109/ICASSP49660.2025.10889000},
  bibsource    = {dblp computer science bibliography, https://dblp.org}
}

@inproceedings{QUEST,
  author       = {Jiaming Tang and
                  Yilong Zhao and
                  Kan Zhu and
                  Guangxuan Xiao and
                  Baris Kasikci and
                  Song Han},
  title        = {{QUEST:} Query-Aware Sparsity for Efficient Long-Context {LLM} Inference},
  booktitle    = {Forty-first International Conference on Machine Learning, {ICML} 2024,
                  Vienna, Austria, July 21-27, 2024},
  publisher    = {OpenReview.net},
  year         = {2024},
  bibsource    = {dblp computer science bibliography, https://dblp.org}
}

@inproceedings{FlexGen,
  author       = {Ying Sheng and
                  Lianmin Zheng and
                  Binhang Yuan and
                  Zhuohan Li and
                  Max Ryabinin and
                  Beidi Chen and
                  Percy Liang and
                  Christopher R{\'{e}} and
                  Ion Stoica and
                  Ce Zhang},
  editor       = {Andreas Krause and
                  Emma Brunskill and
                  Kyunghyun Cho and
                  Barbara Engelhardt and
                  Sivan Sabato and
                  Jonathan Scarlett},
  title        = {FlexGen: High-Throughput Generative Inference of Large Language Models
                  with a Single {GPU}},
  booktitle    = {International Conference on Machine Learning, {ICML} 2023, 23-29 July
                  2023, Honolulu, Hawaii, {USA}},
  series       = {Proceedings of Machine Learning Research},
  volume       = {202},
  pages        = {31094--31116},
  publisher    = {{PMLR}},
  year         = {2023},
  bibsource    = {dblp computer science bibliography, https://dblp.org}
}

@inproceedings{QuaRot,
  author       = {Saleh Ashkboos and
                  Amirkeivan Mohtashami and
                  Maximilian L. Croci and
                  Bo Li and
                  Pashmina Cameron and
                  Martin Jaggi and
                  Dan Alistarh and
                  Torsten Hoefler and
                  James Hensman},
  editor       = {Amir Globersons and
                  Lester Mackey and
                  Danielle Belgrave and
                  Angela Fan and
                  Ulrich Paquet and
                  Jakub M. Tomczak and
                  Cheng Zhang},
  title        = {QuaRot: Outlier-Free 4-Bit Inference in Rotated LLMs},
  booktitle    = {Advances in Neural Information Processing Systems 38: Annual Conference
                  on Neural Information Processing Systems 2024, NeurIPS 2024, Vancouver,
                  BC, Canada, December 10 - 15, 2024},
  year         = {2024},
  bibsource    = {dblp computer science bibliography, https://dblp.org}
}

@article{SKVQ,
  author       = {Haojie Duanmu and
                  Zhihang Yuan and
                  Xiuhong Li and
                  Jiangfei Duan and
                  Xingcheng Zhang and
                  Dahua Lin},
  title        = {{SKVQ:} Sliding-window Key and Value Cache Quantization for Large
                  Language Models},
  journal      = {CoRR},
  volume       = {abs/2405.06219},
  year         = {2024},
  doi          = {10.48550/ARXIV.2405.06219},
  eprinttype    = {arXiv},
  eprint       = {2405.06219},
  bibsource    = {dblp computer science bibliography, https://dblp.org}
}

@inproceedings{AsymKV,
  author       = {Qian Tao and
                  Wenyuan Yu and
                  Jingren Zhou},
  editor       = {Owen Rambow and
                  Leo Wanner and
                  Marianna Apidianaki and
                  Hend Al{-}Khalifa and
                  Barbara Di Eugenio and
                  Steven Schockaert},
  title        = {AsymKV: Enabling 1-Bit Quantization of {KV} Cache with Layer-Wise
                  Asymmetric Quantization Configurations},
  booktitle    = {Proceedings of the 31st International Conference on Computational
                  Linguistics, {COLING} 2025, Abu Dhabi, UAE, January 19-24, 2025},
  pages        = {2316--2328},
  publisher    = {Association for Computational Linguistics},
  year         = {2025},
  bibsource    = {dblp computer science bibliography, https://dblp.org}
}

@inproceedings{ZipCache,
  author       = {Yefei He and
                  Luoming Zhang and
                  Weijia Wu and
                  Jing Liu and
                  Hong Zhou and
                  Bohan Zhuang},
  editor       = {Amir Globersons and
                  Lester Mackey and
                  Danielle Belgrave and
                  Angela Fan and
                  Ulrich Paquet and
                  Jakub M. Tomczak and
                  Cheng Zhang},
  title        = {ZipCache: Accurate and Efficient {KV} Cache Quantization with Salient
                  Token Identification},
  booktitle    = {Advances in Neural Information Processing Systems 38: Annual Conference
                  on Neural Information Processing Systems 2024, NeurIPS 2024, Vancouver,
                  BC, Canada, December 10 - 15, 2024},
  year         = {2024},
  bibsource    = {dblp computer science bibliography, https://dblp.org}
}

@inproceedings{CQ,
  author       = {Tianyi Zhang and
                  Jonah Yi and
                  Zhaozhuo Xu and
                  Anshumali Shrivastava},
  editor       = {Amir Globersons and
                  Lester Mackey and
                  Danielle Belgrave and
                  Angela Fan and
                  Ulrich Paquet and
                  Jakub M. Tomczak and
                  Cheng Zhang},
  title        = {{KV} Cache is 1 Bit Per Channel: Efficient Large Language Model Inference
                  with Coupled Quantization},
  booktitle    = {Advances in Neural Information Processing Systems 38: Annual Conference
                  on Neural Information Processing Systems 2024, NeurIPS 2024, Vancouver,
                  BC, Canada, December 10 - 15, 2024},
  year         = {2024},
  bibsource    = {dblp computer science bibliography, https://dblp.org}
}

@inproceedings{RVQ,
  author       = {Ankur Kumar},
  editor       = {Mehdi Rezagholizadeh and
                  Peyman Passban and
                  Soheila Samiee and
                  Vahid Partovi Nia and
                  Yu Cheng and
                  Yue Deng and
                  Qun Liu and
                  Boxing Chen},
  title        = {Residual vector quantization for {KV} cache compression in large language
                  model},
  booktitle    = {NeurIPS Efficient Natural Language and Speech Processing Workshop,
                  14 December 2024, Vancouver, British Columbia, Canada},
  series       = {Proceedings of Machine Learning Research},
  volume       = {262},
  pages        = {485--490},
  publisher    = {{PMLR}},
  year         = {2024},
  bibsource    = {dblp computer science bibliography, https://dblp.org}
}

@article{Quantize_What_Counts,
  title={Quantize What Counts: Bit Allocation Insights Informed by Spectral Gaps in Keys and Values},
  author={Hariri, Mohsen and Luo, Alan and Nemati, Mohammadreza and Nguyen, Lam and Zhong, Shaochen and Wang, Qifan and Hu, Xia and Han, Xiaotian and Chaudhary, Vipin},
  journal={arXiv preprint arXiv:2502.15075},
  year={2025}
}
\bibliographystyle{icml2026}

\newpage
\appendix
\onecolumn
\section{K Cache Pattern Stable Analysis}
\label{app:k_stable_analysis}
Prior studies have largely focused on outliers in K cache along a single trajectory, with limited evaluation of cross-trajectory consistency under different sampling paradigms. To address this, we build an evaluation set from GSM8K and run the model under two settings: parallel inference and multi-sample decoding. We then compute and compare mutual information for three cases: between tokens across different prefill runs, between tokens across distinct inference trajectories, and between different token positions within a single trajectory. Higher mutual information indicates greater common structure in K cache and stronger consistency, both across and within trajectories.

\begin{table}[ht]
\caption{Mutual Information of K Across Prefill Runs, Trajectories, and Token Positions}
\label{tab:mi}
\small
\renewcommand{\arraystretch}{1.15} 
\begin{center}
\begin{tabular}{lcccc}
\toprule
\multicolumn{1}{c}{\bf Model} &
\multicolumn{1}{c}{\bf Random} &
\multicolumn{1}{c}{\bf Inter-Prefill} &
\multicolumn{1}{c}{\bf Inter-Sample} &
\multicolumn{1}{c}{\bf Inter-Token} \\
\midrule
Llama-3.1-8B-Instruct & \multirow{3}{*}{0.0039} & 0.1868 & 0.1771 & 0.1829 \\
Mistral-7B-Instruct-v0.3  & & 0.2067 & 0.2224 & 0.2291 \\
Qwen2.5-7B-Instruct & & 0.4169 & 0.4169 & 0.4121 \\
\bottomrule
\end{tabular}
\end{center}
\end{table}
From table \ref{tab:mi}, mutual information measured on the K-cache differs across model families; however, for any fixed model, the K-cache mutual information remains highly consistent across settings. Since the primary variation across inference paradigms lies in the composition of the presented context and the resulting trajectories, we arrive at the following observation: For any context, a given model’s K-cache retains a nontrivial amount of stable structural information.

\section{Supplementary Figures For Insight 1 And Insight 2}
\label{app:more_figs}

\paragraph{Settings} All preliminary experiments were conducted on the GSM8K dataset. To ensure fair comparison, we employed a unified zero-shot chain-of-thought (CoT) prompt template across all models and experimental conditions. This design eliminates variability due to prompt formatting and allows a cleaner assessment of the impact of the proposed method.

\begin{AIbox}{CoT Prompt}
    \{Question\}Please reason step by step, and put your final answer within \textbackslash boxed\{\}.
\end{AIbox}

\begin{figure}[h!]
\centering
\includegraphics[width=.9\textwidth]{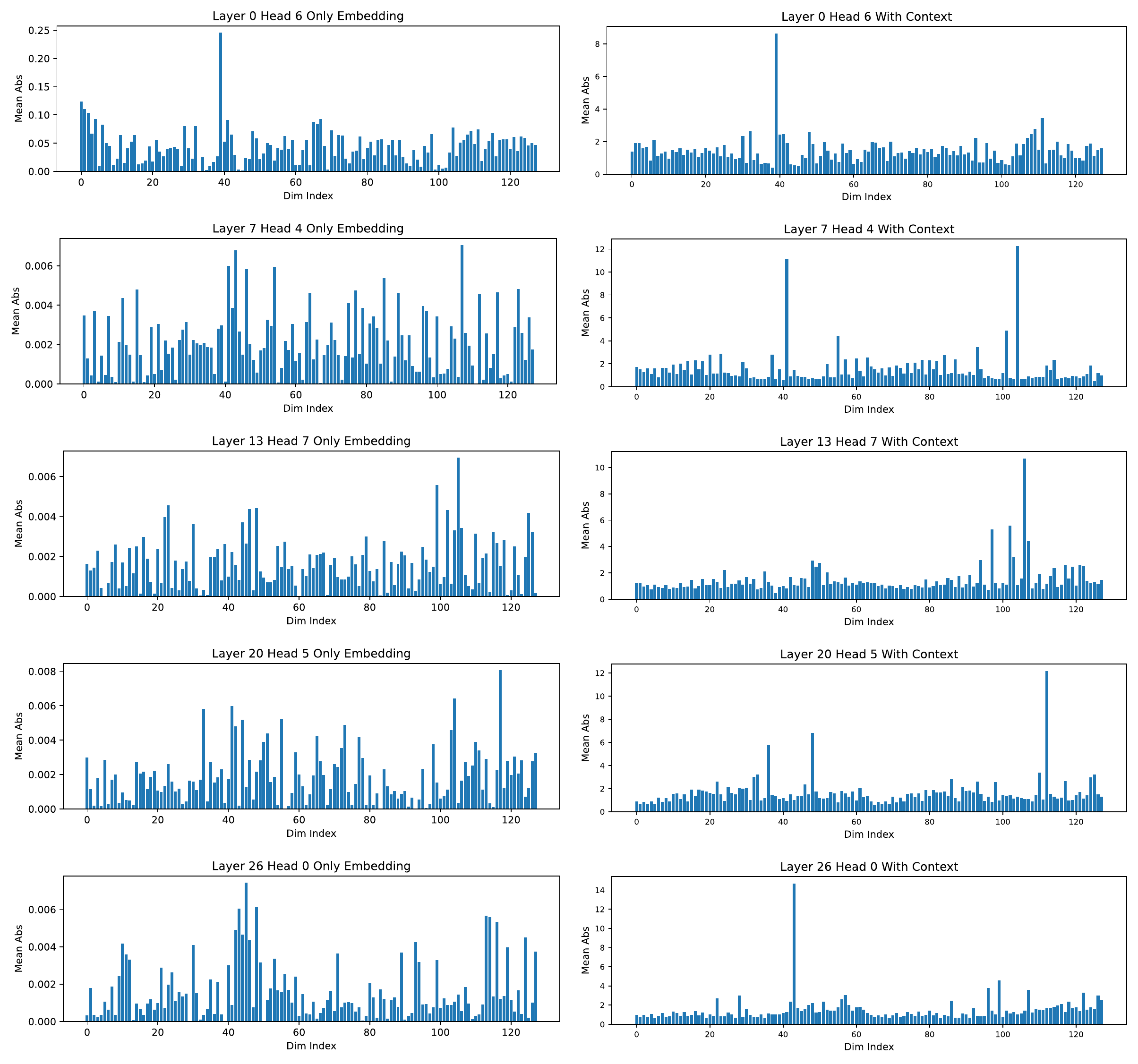}
\caption{\textbf{Additional evidence for Insight 1.} We observe similar phenomena across different layers, supporting that the K-cache stable structure chiefly originates from the model.}
\label{fig:Insight1_extend}
\end{figure}

\begin{figure}[h!]
\centering
\includegraphics[width=.8\textwidth]{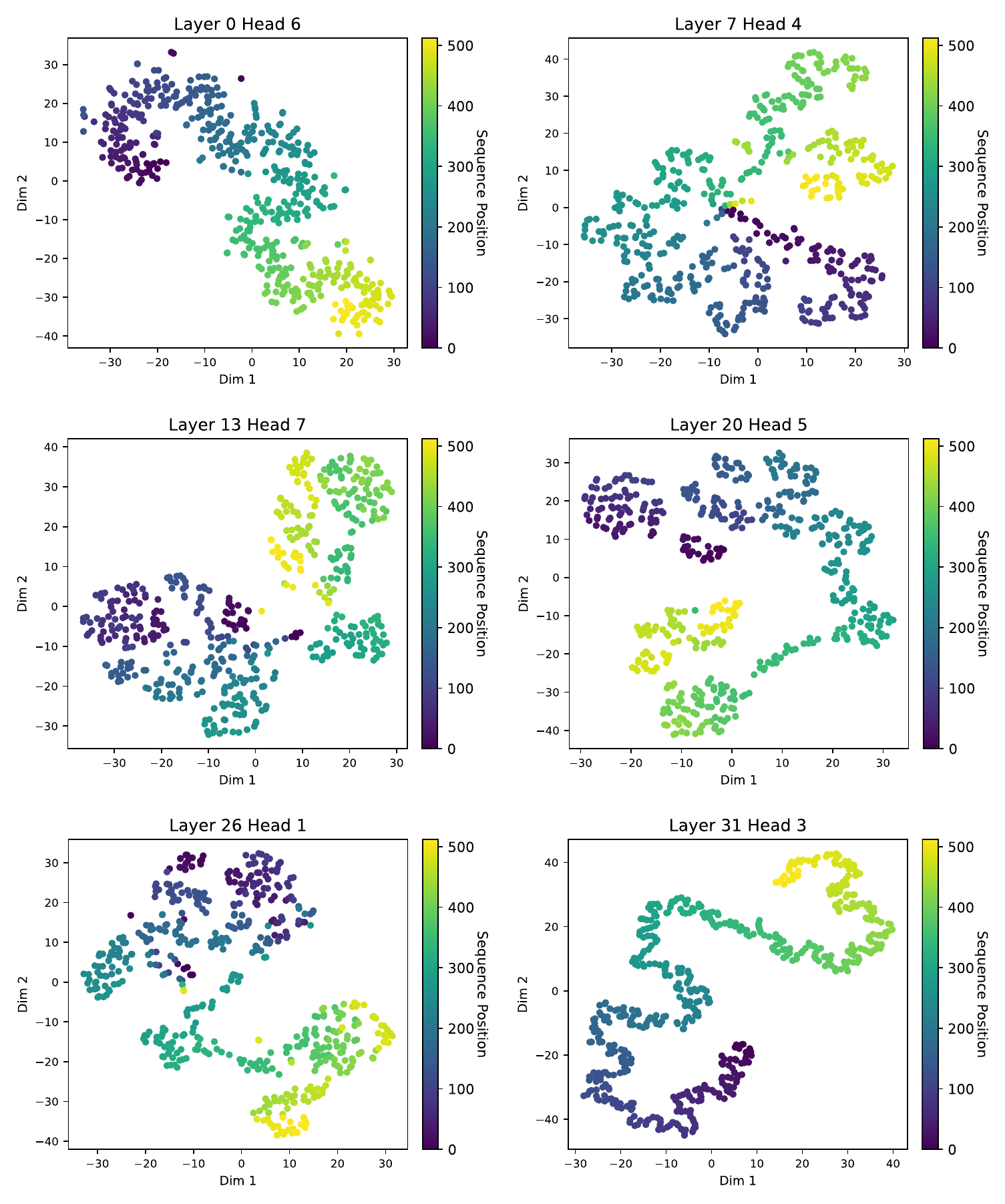}
\caption{\textbf{Additional evidence for Insight 2.} The K cache shows layer- and head-specific evolution as the context grows over the decoding trajectory.}
\label{fig:Insight2_extend}
\end{figure}

We provide additional experimental observations that corroborate our insights. See Figs.~\ref{fig:Insight1_extend} and \ref{fig:Insight2_extend}.


\newpage

\section{Additional Proof}
\label{app:prove_worst_case}
\noindent\textbf{Goal} Show that for any bit-width $b\ge 1$ and any $\rho\in(0,1)$, there exists a finite pattern set $\mathcal P$ such that the residual scheme attains a uniform worst-case error bound satisfying
$$
U_{\mathrm{res}}^\star(b)\ \le\ \rho\,U_{\mathrm{raw}}^\star(b).
$$
and this guarantee holds independently of the sequence length.

Let $X\in\{K,V\}$ denote the per-token key or value in $\mathbb{R}^d$. For any model input and position $t\ge 1$, write $X_t$ for the resulting vector.
Assume bounded token embeddings and positional signals:
\begin{align}
\max_{w}\|\mathrm{emb}(w)\|_2\le M,
\qquad
\sup_{t\ge 1}\|\mathrm{pos\_emb}(t)\|_2\le N.
\end{align}
Let $H^{(\ell)}_t$ be the hidden state at layer $\ell$. Denote by $\mathrm{LN}$ the normalization used in the block and by $\Psi$ the remainder of the block’s mapping (attention + FFN + residual, etc.). We assume:
\begin{align}
\|\Psi(h)\|_2 \le L_\Psi\|h\|_2 + B_\Psi,\qquad
\|\mathrm{LN}(h)\|_2 \le c_{\mathrm{LN}}\|h\|_2 + b_{\mathrm{LN}}
\end{align}
for constants $L_\Psi,B_\Psi,c_{\mathrm{LN}},b_{\mathrm{LN}}$ that do not depend on sequence length or position. For the head’s projection to $X$, write
\begin{align}
X = W_X\,\mathrm{LN}\bigl(H^{(\ell)}\bigr),
\qquad \|W_X\|_{2\to2} = \sigma_X.
\end{align}

The input satisfies $H^{(0)}_t=\mathrm{emb}(w_t)+\mathrm{pos\_emb}(t)$ and hence $\sup_t\|H^{(0)}_t\|_2\le M+N$. Define $S_\ell:=\sup_{t\ge 1}\|H^{(\ell)}_t\|_2$. Using the residual update and the linear hypotheses,
\begin{align}
\|H^{(\ell+1)}_t\|_2
\le \|H^{(\ell)}_t\|_2 + \|\Psi(\mathrm{LN}(H^{(\ell)}_t))\|_2
\le a\,\|H^{(\ell)}_t\|_2 + b,
\end{align}
where $a:=1+L_\Psi c_{\mathrm{LN}}$ and $b:=L_\Psi b_{\mathrm{LN}}+B_\Psi$. Taking suprema over $t$ gives
\begin{align}
S_{\ell+1}\le a\,S_\ell + b,
\qquad
S_0\le M+N
\quad\Rightarrow\quad
S_\ell\le a^\ell(M+N)+\frac{a^\ell-1}{a-1}\,b.
\end{align}
Therefore,
\begin{align}
\sup_{t\ge 1}\|X_t\|_2
=\sup_t\|W_X\,\mathrm{LN}(H^{(\ell)}_t)\|_2
\le \sigma_X\bigl(c_{\mathrm{LN}}S_\ell+b_{\mathrm{LN}}\bigr)
=: R_2<\infty.
\end{align}
Let
\begin{align}
\mathcal{S}_X:=\{X_t:\ \text{all inputs, all }t\ge 1\}\subset B_2(0,R_2)\subset\mathbb{R}^d.
\end{align}
In finite dimensions, bounded sets are totally bounded: for every $\varepsilon>0$ there exists a finite $\varepsilon$-net $\mathcal N_\varepsilon$ in $\ell_\infty$ such that $\mathcal{S}_X\subset\bigcup_{p\in\mathcal N_\varepsilon}B_\infty(p,\varepsilon)$. Writing $R_\infty:=\sup_{x\in\mathcal{S}_X}\|x\|_\infty\le R_2$, a crude covering estimate is
\begin{align}
N_\infty(\mathcal{S}_X,\varepsilon)\le \bigl(1+2R_\infty/\varepsilon\bigr)^d.
\end{align}

Define the (standard) $\ell_\infty$ Chebyshev radius and center
\begin{align}
R^\star:=\inf_{c\in\mathbb{R}^d}\ \sup_{x\in\mathcal{S}_X}\|x-c\|_\infty,
\qquad
c^\star\in\arg\min_c\sup_{x\in\mathcal{S}_X}\|x-c\|_\infty,
\end{align}
and note that $w(x-c):=\max_i(x_i-c_i)-\min_i(x_i-c_i)\le 2\|x-c\|_\infty$.

To compare worst-case bounds without unnecessary slack, introduce the \emph{width–Chebyshev radius}
\begin{align}
R_w^\star \;:=\; \tfrac12\,\inf_{c\in\mathbb{R}^d}\ \sup_{x\in\mathcal{S}_X} w(x-c),
\qquad
c_w^\star \in \underset{c}{argmin}\ \sup_{x\in\mathcal{S}_X} w(x-c).
\end{align}
For non-symmetric uniform min–max quantization on a group of size $g$ and $b$ bits, the \emph{optimal uniform worst-case bound (OUWB)} for the \emph{direct} scheme is
\begin{align}
U_{\mathrm{raw}}^\star(b)\;:=\;\inf_{c}\ \sup_{x\in\mathcal{S}_X}\ \frac{\sqrt{g}}{2}\,\frac{w(x-c)}{2^b-1}
\;=\;\frac{\sqrt{g}}{2}\,\frac{2R_w^\star}{2^b-1}.
\end{align}

Let $w(z):=\max_i z_i-\min_i z_i$. Fix any $\rho\in(0,1)$ and set $\varepsilon=\rho R_w^\star$. By total boundedness, select a finite $\varepsilon$-net $\mathcal P=\{p_1,\dots,p_K\}$ in $\ell_\infty$ covering $\mathcal{S}_X$. For any $x\in\mathcal{S}_X$, choose $p(x)\in\mathcal P$ with $\|x-p(x)\|_\infty\le \varepsilon$. Then
\begin{align}
w\bigl(x-p(x)\bigr) \le 2\|x-p(x)\|_\infty \le 2\rho R_w^\star,
\qquad
\text{hence}\quad
\sup_x \frac{\sqrt{g}}{2}\frac{w(x-p(x))}{2^b-1}
\ \le\ \rho\,U_{\mathrm{raw}}^\star(b).
\end{align}
Infimizing over finite $\mathcal P$ yields the residual OUWB
\begin{align}
U_{\mathrm{res}}^\star(b)\ \le\ \rho\,U_{\mathrm{raw}}^\star(b).
\end{align}

Consequently, for any $b\ge 1$ and $\rho\in(0,1)$, there exists a finite pattern set $\mathcal P$ such that the residual scheme achieves an optimal uniform worst-case bound that is a $\rho$-fraction of the direct scheme’s optimal uniform worst-case bound, independently of sequence length.

\section{Detailed Information of LongBench}
\label{app:longbench}
Following the LongBench official documentation, we categorize tasks into six types. The tasks and accompanying configurations for each category are listed in Table \ref{tab:longhbench_detail}.
\begin{table}[ht]
\caption{LongBench Overview}
\label{tab:longhbench_detail}
\small
\renewcommand{\arraystretch}{1.15} 
\begin{center}
\begin{tabular}{c l l r c r}
\toprule
\multicolumn{1}{c}{\bf Task Type} &
\multicolumn{1}{c}{\bf Task} &
\multicolumn{1}{c}{\bf Metric} &
\multicolumn{1}{c}{\bf Avg. Length} &
\multicolumn{1}{c}{\bf Language} &
\multicolumn{1}{c}{\bf \#Samples} \\
\toprule

\multirow{5}{*}{Multi-document QA}
  & HotpotQA           & F1       & 9151  & English & 200 \\
  & 2WikiMultihopQA    & F1       & 4887  & English & 200 \\
  & MuSiQue            & F1       & 11214 & English & 200 \\
  & DuReader           & Rouge-L  & 15768 & Chinese & 200 \\
  & MultiFieldQA-zh    & F1       & 6701  & Chinese & 200 \\
\midrule
\multirow{3}{*}{Single-document QA}
  & MultiFieldQA-en    & F1       & 4559  & English & 150 \\
  & NarrativeQA        & F1       & 18409 & English & 200 \\
  & Qasper             & F1       & 3619  & English & 200 \\
\hline
\multirow{4}{*}{Summarization}
  & GovReport          & Rouge-L  & 8734  & English & 200 \\
  & QMSum              & Rouge-L  & 10614 & English & 200 \\
  & MultiNews          & Rouge-L  & 2113  & English & 200 \\
  & VCSUM              & Rouge-L  & 15380 & Chinese & 200 \\
\midrule
\multirow{4}{*}{Few-shot}
  & TriviaQA           & F1       & 8209  & English & 200 \\
  & SAMSum             & Rouge-L  & 6258  & English & 200 \\
  & TREC               & Accuracy & 5177  & English & 200 \\
  & LSHT               & Accuracy & 22337 & Chinese & 200 \\
\midrule
\multirow{3}{*}{Synthetic Task}
  & PassageRetrieval-en & Accuracy & 9289  & English & 200 \\
  & PassageCount        & Accuracy & 11141 & English & 200 \\
  & PassageRetrieval-zh & Accuracy & 6745  & Chinese & 200 \\
\midrule
\multirow{2}{*}{Code}
  & LCC                & Edit Sim & 1235  & Python/C\#/Java & 500 \\
  & RepoBench-P        & Edit Sim & 4206  & Python/Java     & 500 \\
\bottomrule
\end{tabular}
\end{center}
\end{table}

\section{Baseline Settings}
\label{app:baseline}

This section details the baseline configurations. For KIVI~\citep{KIVI}, we set $\text{group\_size}=128$ and $\text{residual\_size}=128$. For ZipCache~\citep{ZipCache}, we assign $\text{unimportant\_ratio}=0.875$ to both the K and V caches to approximately align the memory footprint. For SKVQ~\citep{SKVQ}, we follow the official implementation with $\text{group\_size}=128$, channel-reorder count of 8, and $\text{clip\_ratio}=0.92$. For OTT~\citep{OTT}, we configure $\text{group\_size}=128$, $\text{residual\_size}=32$, $\text{sink\_num}=3$, and $\text{max\_sink\_num}=32$.

A potential source of confusion is the fact that the residual size differs between KIVI and OTT. This discrepancy does not bias the comparison; on the contrary, it is introduced to make the comparison fairer. In KIVI, the quantization strategy on the K cache does not maintain a continuous sliding window: after each group is quantized, that group is immediately cleared from the residual buffer. In contrast, OTT maintains a fixed sliding window of length $\text{residual\_size}$ throughout the entire decoding process, so that the last $\text{residual\_size}$ tokens always remain in higher precision.

To keep the effective number of FP16 tokens preserved during decoding aligned across methods, we therefore set OTT’s $\text{residual\_size}=32$. This choice ensures that the number of unquantized tokens retained by OTT at each decoding step is comparable, on average, to the number of FP16 tokens maintained by KIVI under its group-based quantization behavior.

\section{INT4 Results on LongBench}

In the 4-bit setting, we evaluate our method alongside baselines. As shown in Table~\ref{tab:longbench-results-int4}, our method incurs only a 0.08\% accuracy drop relative to FP16, which is nearly lossless.
\label{app:longbench_int4}
\begin{table}[ht]
\caption{Overall results on LongBench at 4-bit setting. The best and second-best in every column are marked in \textbf{bold} and \underline{underline}, respectively.}
\label{tab:longbench-results-int4}
\small
\renewcommand{\arraystretch}{1.15} 
\begin{center}
\begin{tabular}{p{3cm}p{1.2cm}ccccccc}
\toprule
\multicolumn{1}{c}{\bf Model} &
\multicolumn{1}{c}{\bf Method} &
\multicolumn{1}{c}{\bf MQA} &
\multicolumn{1}{c}{\bf SQA} &
\multicolumn{1}{c}{\bf Summ.} &
\multicolumn{1}{c}{\bf Few-shot} &
\multicolumn{1}{c}{\bf Synth.} &
\multicolumn{1}{c}{\bf Code} &
\multicolumn{1}{c}{\bf Avg} \\
\toprule

\multirow{6}{*}{Llama3.1-8B-Instruct}
& FP16 & 36.63 & 46.56 & 25.54 & 61.16 & 59.99 & 59.42 & 46.59 \\
\cmidrule(lr){2-9}
& KIVI       & \underline{36.63} & \textbf{46.69} & \underline{25.64} & \underline{61.25} & 57.77 & 59.48 & \underline{46.34} \\
& ZipCache   & - & - & - & - & - & - & -\\
& SKVQ       & 35.39 & 44.15 & 25.23 & 59.70 & \underline{58.46} & \textbf{63.79} & 45.75\\
& OTT        & 35.39 & 44.61 & \textbf{25.70} & 60.00 & \textbf{58.92} & \underline{63.75} & 46.05 \\
& PatternKV  & \textbf{36.78} & \underline{46.59} & 25.50 & \textbf{61.29} & 58.42 & 59.31 & \textbf{46.41} \\
\midrule

\multirow{6}{*}{Llama3.1-70B-Instruct}
& FP16 & 52.68 & 49.56 & 25.67 & 66.18 & 72.67 & 46.80 & 51.81 \\
\cmidrule(lr){2-9}
& KIVI       & \textbf{53.09} & \underline{49.58} & \underline{25.68} & \textbf{66.16} & \underline{72.67} & 46.80 & \textbf{51.89} \\
& ZipCache   & - & - & - & - & - & - & -\\
& SKVQ       & - & - & - & - & - & - & -\\
& OTT        & 43.17 & 47.96 & 25.10 & 61.01 & 68.67 & \textbf{60.78} & 49.36 \\
& PatternKV  & \underline{52.66} & \textbf{49.70} & \textbf{25.80} & \underline{66.12} & \textbf{72.83} & \underline{46.86} & \underline{51.87} \\
\midrule

\multirow{3}{*}{Qwen2.5-7B-instruct}
& FP16 & 38.03 & 45.40 & 23.37 & 59.85 & 58.83 & 62.84 & 46.13 \\
\cmidrule(lr){2-9}
& KIVI       & \underline{37.71} & \textbf{45.58} & \textbf{23.46} & \underline{59.88} & \underline{58.50} & \underline{62.53} & \underline{46.05} \\
& PatternKV  & \textbf{38.33} & \underline{45.00} & \underline{23.36} & \textbf{60.04} & \textbf{59.17} & \textbf{62.78} & \textbf{46.19}\\
\bottomrule
\end{tabular}
\end{center}
\end{table}

\section{INT4 Results on Long-CoT settings}
\label{app:longcot_int4}
In the 4-bit setting, we evaluate our method against baselines; the results are shown in Table~\ref{tab:long-cot-results-int4}. Overall accuracy is substantially restored, although a residual gap remains. On benchmarks with larger degradation (e.g., AIME25), our method often recovers a substantial portion of the accuracy.

\begin{table}[h!]
\caption{Overall results on long-CoT Benchmark at 4-bit setting.}
\label{tab:long-cot-results-int4}
\small
\renewcommand{\arraystretch}{1.15} 
\setlength{\tabcolsep}{3pt}
\begin{center}
\begin{tabular}{lllcccccccccc}
\toprule
\multicolumn{1}{c}{\bf Model} &
\multicolumn{1}{c}{\bf Method} &
\multicolumn{2}{c}{\bf AIME 25} &
\multicolumn{2}{c}{\bf AIME 24} &
\multicolumn{2}{c}{\bf AMC 24} &
\multicolumn{2}{c}{\bf AMC 23} \\

\cline{3-10}
&  & \multicolumn{1}{c}{Avg@8} & \multicolumn{1}{c}{Maj@8}
 & \multicolumn{1}{c}{Avg@8} & \multicolumn{1}{c}{Maj@8}
 & \multicolumn{1}{c}{Avg@8} & \multicolumn{1}{c}{Maj@8}
 & \multicolumn{1}{c}{Avg@8} & \multicolumn{1}{c}{Maj@8} \\
\toprule

\multirow{3}{*}{Llama-8B}
  & FP16
  & 32.33 & 37.93
  & 37.93 & 61.55
  & 53.06 & 60.22
  & 85.58 & 90.13 \\
\cmidrule(lr){2-10}
& KIVI
  & 24.58 & 31.33
  & 37.92 & 57.16
  & 52.50 & 65.88
  & 86.86 & 92.56 \\
& PatternKV
  & \textbf{27.50} & \textbf{37.0}
  & \textbf{38.75} & \textbf{59.33}
  & 50.83 & 58.22
  & 85.31 & 91.13 \\
\midrule

\multirow{3}{*}{Qwen-7B}
  & FP16
  & 38.39 & 52.14
  & 51.67 & 71.67
  & 60.51 & 63.18
  & 90.06 & 94.87 \\
\cmidrule(lr){2-10}
& KIVI
  & 38.67 & 49.33
  & 49.58 & 73.33
  & 58.89 & 66.0
  & 89.06 & 93.75 \\
& PatternKV
  & 38.33 & 46.33
  & \textbf{52.08} & \textbf{69.5}
  & \textbf{60.28} & \textbf{68.0}
  & 88.44 & \textbf{95.0} \\
\midrule

\multirow{3}{*}{Qwen-14B}
  & FP16
  & 45.83 & 63.17
  & 64.58 & 75.83
  & 65.00 & 67.56
  & 92.50 & 95.0 \\
\cmidrule(lr){2-10}
& KIVI
  & 42.08 & 53.33
  & 63.33 & 76.67
  & 61.67 & 66.89
  & 91.88 & 95.0 \\
& PatternKV
  & \textbf{45.83} & \textbf{64.17}
  & 62.50 & \textbf{76.67}
  & \textbf{61.94} & 64.89
  & \textbf{92.81} & \textbf{95.0}  \\
\bottomrule

\end{tabular}
\end{center}
\end{table}

\newpage

\section{V Pattern Utilization Rate}
\label{app:use_rate}
\begin{figure}[h]
\centering
\includegraphics[width=.9\textwidth]{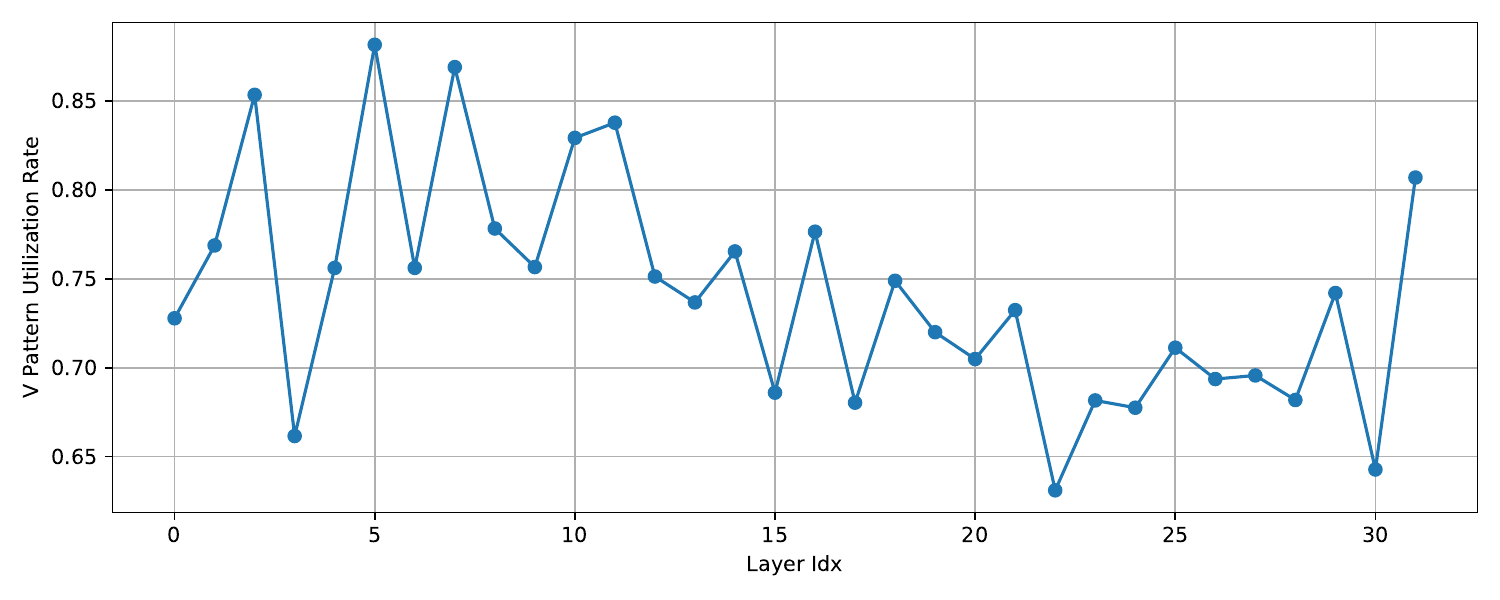}
\caption{Visualization of V Pattern Utilization Rate on \textit{TriviaQA}}
\label{fig:v_use_rate}
\end{figure}

As shown in Fig.~\ref{fig:v_use_rate} for \textit{TriviaQA}, utilization remains high even under thresholding, implying the presence of latent semantic regularities in V cache.

\newpage

\section{Detailed Efficiency Analysis of PatternKV}
\label{app:efficiency-patternkv}

In this section, we provide a component-level breakdown of the computational overhead introduced by PatternKV, together with head-to-head comparisons against KIVI, ZipCache, OTT, and SKVQ in terms of latency, throughput, and peak GPU memory.

\paragraph{Settings} We profile inference throughput and peak memory usage on an NVIDIA H20 (96 GB) GPU. The input and output lengths are fixed at 1024 and 256 tokens, respectively, and we evaluate batch sizes of $\{16, 32, 48, 64, 96, 128, 160\}$. The model is Llama-3.1-8B-Instruct.

\subsection{Component-Level Breakdown}

The additional computational components of PatternKV, beyond standard KV quantization, are:

\begin{itemize}
    \item \emph{Pattern mining (prefill).} K-means clustering over \(|\mathcal{M}|\) patterns during the prefill stage.
    \item \emph{Chebyshev-center updates (decode).} Non-iterative updates of pattern centers as decoding proceeds.
    \item \emph{Pattern selection (decode).} At each decoding step, selecting the appropriate pattern via an index lookup and gathering the corresponding centroids.
\end{itemize}

\begin{table}[ht]
    \centering
    \caption{Component-level latency, throughput, and FLOPs associated with the PatternKV-specific procedures (batch size \(=32\)). Latency and throughput are reported as relative change with respect to the baseline attention pipeline (negative throughput indicates a drop).}
    \label{tab:patternkv-component-breakdown}
    \begin{tabular}{lccc}
        \toprule
        Component                & Latency  & Throughput & GFLOPs \\
        \midrule
        Pattern mining           & \(+6.60\%\) & \(-6.10\%\) & 6.91 \\
        Pattern selection        & \(+2.59\%\) & \(-2.47\%\) & 2.07 \\
        Chebyshev-center updates & \(+3.11\%\) & \(-2.94\%\) & 2.78 \\
        \bottomrule
    \end{tabular}
\end{table}

Table~\ref{tab:patternkv-component-breakdown} reports the overhead of three PatternKV-specific procedures at batch size \(32\). The dominant source of overhead is the pattern mining step in the prefill stage, which is expected since K-means requires iterative updates to converge. In comparison, pattern selection and Chebyshev-center updates each contribute only about 2\%-3\% additional latency. Overall, these components introduce only modest extra cost relative to the full attention pipeline, while enabling the accuracy gains brought by PatternKV.

\subsection{Head-to-Head Latency and Throughput}

Under identical hardware, model, quantization configuration, batch sizes, and sequence lengths, we compare end-to-end generation latency and throughput across FP16, KIVI, OTT, ZipCache, SKVQ, and PatternKV.

\paragraph{Latency} Table~\ref{tab:patternkv-latency} summarizes the per-token latency (in seconds) for different batch sizes.

\begin{table}[t]
    \centering
    \caption{End-to-end latency (s) under different batch sizes. ``--'' indicates that the corresponding configuration does not fit in GPU memory.}
    \label{tab:patternkv-latency}
    \begin{tabular}{lrrrrrrr}
        \toprule
        Method    & bz\(=16\) & bz\(=32\) & bz\(=48\) & bz\(=64\) & bz\(=96\) & bz\(=128\) & bz\(=160\) \\
        \midrule
        FP16      & 11.29 & 18.39 & 24.46 & 30.57 & 45.78 & 58.70 & --    \\
        KIVI      & 11.48 & 12.10 & 13.37 & 15.13 & 22.29 & 27.30 & 36.70 \\
        OTT       & 10.34 & 12.72 & 15.23 & 17.59 & 25.93 & 31.96 & --    \\
        ZipCache  & 23.02 & 40.84 & 57.27 & 74.04 & 109.89 & 143.13 & --   \\
        SKVQ      & 16.73 & 26.87 & 77.95 & 182.45 & 220.83 & --   & --    \\
        PatternKV &  9.80 & 13.49 & 16.73 & 19.86 & 29.36 & 36.69 & 48.37 \\
        \bottomrule
    \end{tabular}
\end{table}

\paragraph{Throughput} Table~\ref{tab:patternkv-throughput} reports the corresponding throughput (tokens per second).

\begin{table}[t]
    \centering
    \caption{End-to-end throughput (tokens/s) under different batch sizes. ``--'' indicates that the corresponding configuration does not fit in GPU memory.}
    \label{tab:patternkv-throughput}
    \begin{tabular}{lrrrrrrr}
        \toprule
        Method    & bz\(=16\) & bz\(=32\) & bz\(=48\) & bz\(=64\) & bz\(=96\) & bz\(=128\) & bz\(=160\) \\
        \midrule
        FP16      & 1089.33 & 1337.54 & 1508.60 & 1609.53 & 1612.52 & 1676.64 & --    \\
        KIVI      & 1071.59 & 2032.88 & 2759.44 & 3251.66 & 3311.32 & 3605.49 & 3352.05 \\
        OTT       & 1189.93 & 1934.07 & 2423.10 & 2796.90 & 2846.63 & 3079.22 & --    \\
        ZipCache  &  534.36 &  602.51 &  644.50 &  664.68 &  671.78 &  687.67 & --    \\
        SKVQ      &  735.25 &  915.52 &  473.48 &  269.74 &  334.29 & --      & --    \\
        PatternKV & 1255.05 & 1823.69 & 2206.27 & 2477.12 & 2514.12 & 2682.75 & 2543.68 \\
        \bottomrule
    \end{tabular}
\end{table}

From Tables~\ref{tab:patternkv-latency} and~\ref{tab:patternkv-throughput}, we observe that PatternKV incurs an increase in latency and a decrease in throughput compared with KIVI at larger batch sizes. This is expected, as PatternKV introduces additional components into the inference pipeline. At the same time, PatternKV delivers substantial accuracy improvements over KIVI (see main text), leading to a more favorable overall trade-off between efficiency and performance. Compared with ZipCache and SKVQ, PatternKV attains consistently better latency and throughput across all reported batch sizes, and it is competitive with OTT while being more accurate.

\subsection{Peak Memory Overhead}

We next provide a fine-grained analysis of the memory overhead of PatternKV and validate it using empirical peak GPU memory measurements.

\paragraph{Analytical memory estimates}
We maintain a separate set of centroids for each attention head, which introduces additional memory overhead. Let \(L\) denote the number of layers, \(H\) the number of KV heads, \(B\) the batch size, and \(S\) the sequence length (in tokens). For INT2 quantization, each stored value occupies \(2\) bytes.

\emph{Prefill stage} In the prefill stage, the number of pattern vectors is fixed regardless of the context length. The memory footprint of pattern vectors can be approximated as
\[
M_{\mathrm{pattern, prefill}} = 2 L H \cdot 32 \cdot 2 \ \mathrm{bytes}
\]
where the leading factor \(2\) accounts for both \(K\) and \(V\), and \(32\) is the number of pattern centroids per head. The KV cache memory usage is
\[
M_{\mathrm{KV}} = 2 B L H S \cdot 2 \ \mathrm{bytes}
\]
The relative memory fraction of pattern vectors during prefill is therefore
\[
\frac{M_{\mathrm{pattern, prefill}}}{M_{\mathrm{KV}}} = \frac{32}{B S}
\]
For a medium-length sequence with \(S = 8192\) (approximately \(8\mathrm{K}\)) and \(B = 1\), the pattern vectors account for only about \(0.39\%\) of the KV memory. As the batch size increases, this fraction decreases further.

\emph{Decode stage} In the decode stage, we set \(G_{\mathrm{pattern}} = 128\), meaning that one pattern vector is generated for every \(128\) decoding steps. Analogously, the relative memory ratio of pattern vectors during decoding is
\[
\frac{M_{\mathrm{pattern, decode}}}{M_{\mathrm{KV}}} = \frac{1}{B G_{\mathrm{pattern}}}
\]
This ratio is at most \(0.78\%\) when \(B = 1\), and again decreases as the batch size grows. In our implementation, the extra centroid tensors account for only about \(0.42\%\) of the overall KV-cache memory footprint.

\paragraph{Empirical peak GPU memory}
Table~\ref{tab:patternkv-peak-memory} reports the peak GPU memory of each method during inference, which corroborates the above estimates.

\begin{table}[t]
    \centering
    \caption{Peak GPU memory consumption (GB) under different batch sizes.}
    \label{tab:patternkv-peak-memory}
    \begin{tabular}{lrrrrrrr}
        \toprule
        Method    & bz\(=16\) & bz\(=32\) & bz\(=48\) & bz\(=64\) & bz\(=96\) & bz\(=128\) & bz\(=160\) \\
        \midrule
        FP16      & 21.93 & 28.84 & 35.83 & 42.77 & 56.67 & 70.56 & --    \\
        KIVI      & 21.19 & 27.39 & 33.57 & 39.77 & 52.16 & 64.55 & 76.95 \\
        OTT       & 24.07 & 31.13 & 38.15 & 45.20 & 59.31 & 73.41 & --    \\
        ZipCache  & 23.10 & 29.21 & 35.35 & 41.43 & 53.61 & 65.84 & --    \\
        SKVQ      & 24.68 & 32.37 & 40.06 & 47.75 & 63.13 & --    & --    \\
        PatternKV & 21.23 & 27.48 & 33.69 & 39.90 & 52.39 & 64.81 & 77.28 \\
        \bottomrule
    \end{tabular}
\end{table}

Compared with the KIVI baseline, PatternKV introduces only about \(0.42\%\) additional peak memory overhead while yielding a clear improvement in performance. This indicates that a very small extra storage cost is sufficient to obtain meaningful performance gains with PatternKV. In contrast, OTT, ZipCache, and SKVQ exhibit higher peak memory usage and less favorable scaling with the batch dimension, largely due to limitations in the publicly available implementations. Both KIVI and PatternKV maintain a well-controlled peak memory profile across the evaluated batch sizes.

\section{K-drift under ALiBi positional encoding}
\label{app:alibi-k-drift}

\begin{figure}[ht]
    \centering
    \includegraphics[width=0.60\linewidth]{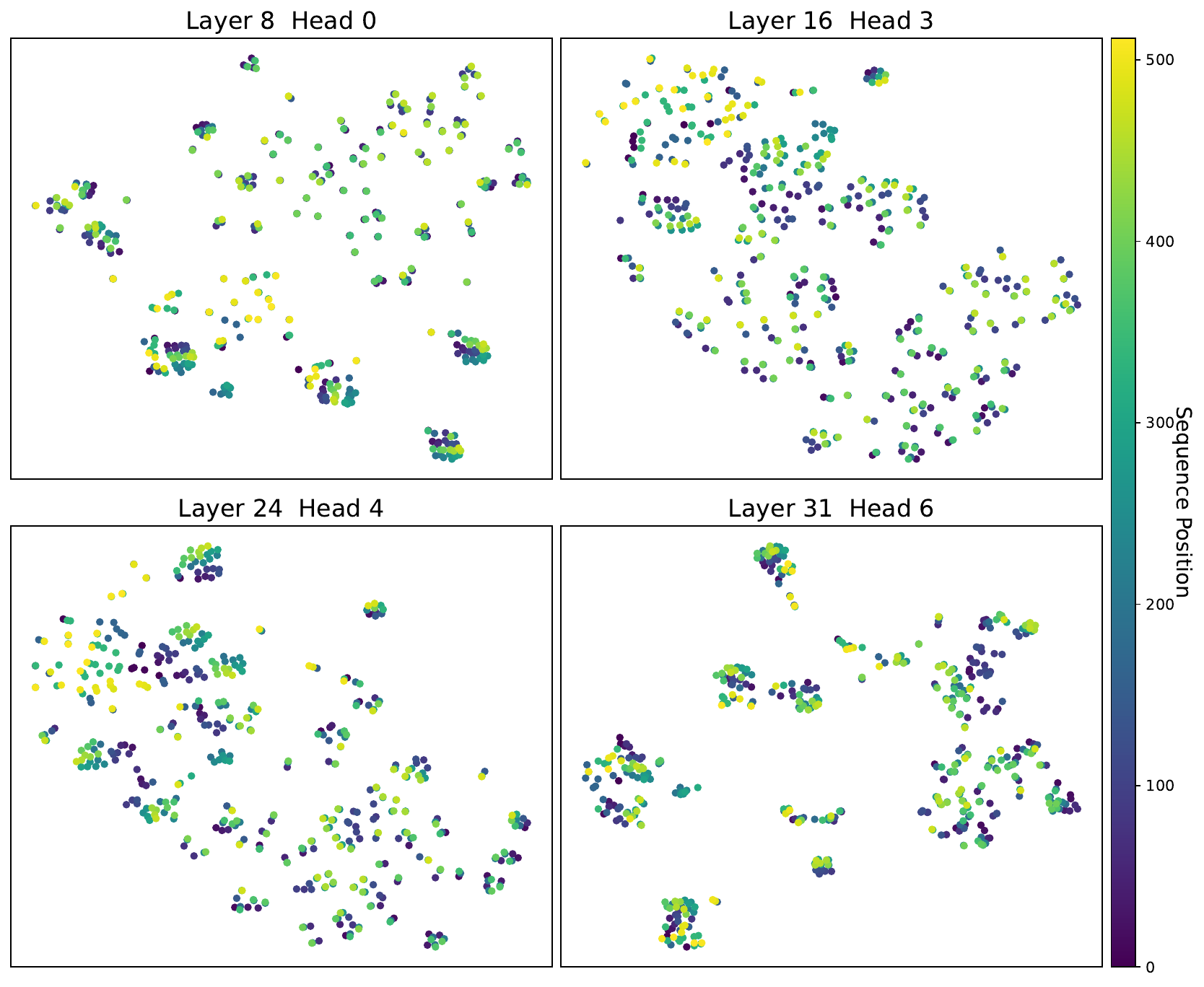}
    \caption{Pattern evolution of the K cache under ALiBi positional encoding on MPT-7B-Chat. Each row corresponds to an attention head, and colors denote discovered K patterns. Tokens from early and late positions can share similar patterns and abrupt pattern changes may occur between adjacent positions, in contrast to the smooth K-drift observed in RoPE-based models (see Figure~\ref{fig:k_evo_analysis_and_v_analysis}).}
    \label{fig:alibi-k-pattern-evolution}
\end{figure}

To further examine whether the observed drift of K patterns is primarily induced by RoPE, we additionally conduct pattern-evolution experiments on MPT-7B-Chat, a decoder-only model that adopts ALiBi positional encoding.
Unlike RoPE-based models, the K cache in MPT-7B-Chat does not exhibit a clear progressive evolution with sequence length: tokens from early and late positions can share similar patterns, and abrupt changes may occur even between adjacent positions.
Figure~\ref{fig:alibi-k-pattern-evolution} visualizes the pattern assignments along the sequence dimension for representative attention heads.
As shown in the figure, the pattern assignments under ALiBi remain relatively mixed across positions, without the smooth positional drift observed in the RoPE-based setting.
These observations provide further evidence that the systematic drift of K patterns reported in the main paper is largely a consequence of RoPE, rather than an inherent property of attention weights or of the proposed pattern-mining procedure.

\newpage

\section{PatternKV under Multi-Head Latent Attention (MLA)}

To assess how PatternKV extends to models with Multi-Head Latent Attention (MLA) and other non-standard attention architectures, we take DeepSeek-V2-Lite-Chat as a representative MLA model. Since, in MLA, RoPE is no longer applied to the entire K cache, we first verify whether the K cache in MLA still exhibits the key properties observed in GQA/MHA models.

\begin{figure}[ht]
    \centering
    \includegraphics[width=0.65\linewidth]{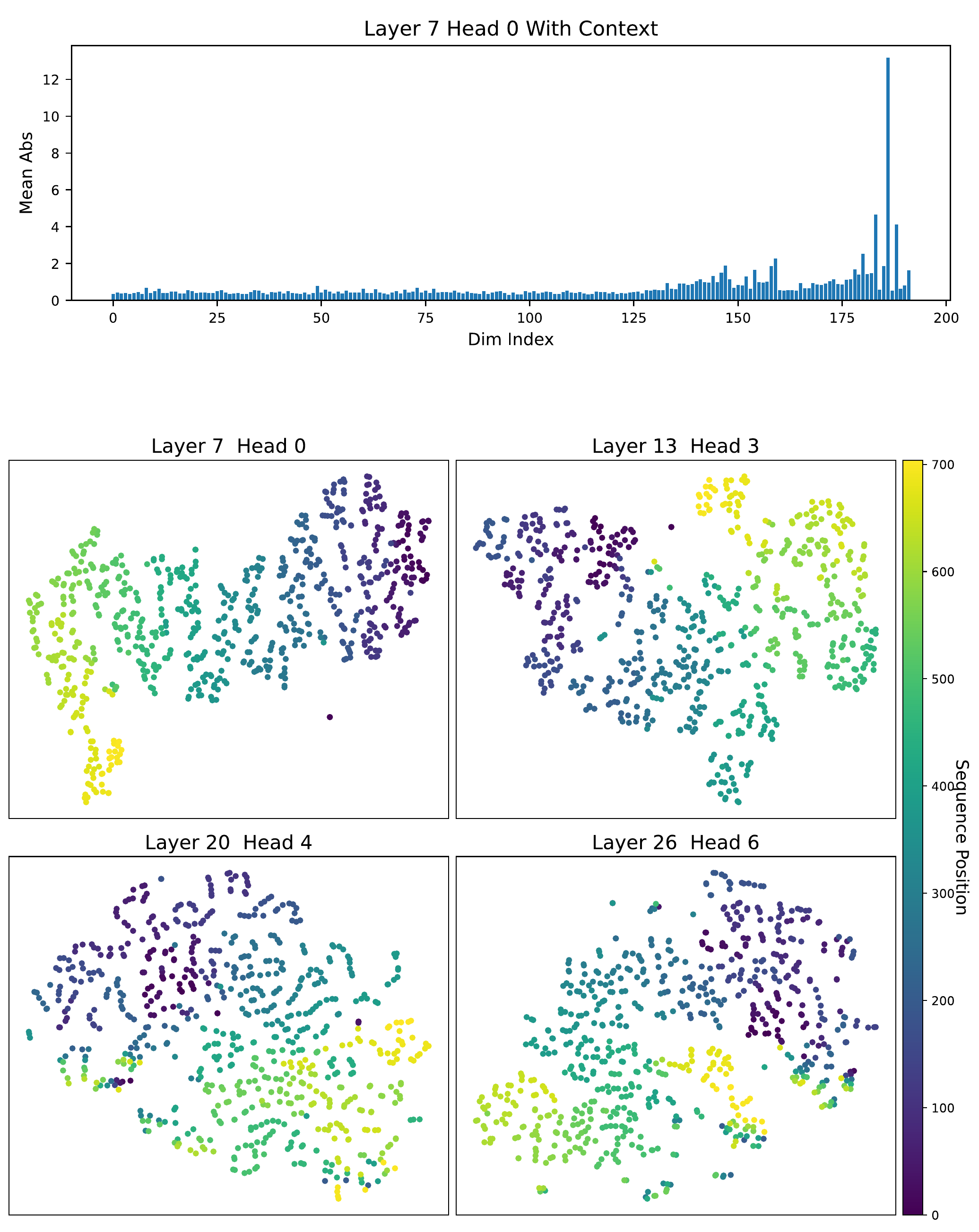}
    \caption{Pattern evolution of the K cache in DeepSeek-V2-Lite-Chat (MLA). The RoPE-applied subspace exhibits larger magnitudes and a gradual evolution along the sequence dimension, while remaining stable along channels, which mirrors the behavior observed in GQA/MHA and supports the applicability of PatternKV to MLA.}
    \label{fig:mla-k-pattern-evolution}
\end{figure}

Based on Figure \ref{fig:mla-k-pattern-evolution} , we make the following observations:
\begin{itemize}
    \item The RoPE-applied part of the K cache in MLA has substantially larger magnitudes than the non-RoPE part, which further suggests that RoPE is the primary source of the outlier distribution in the K cache.
    \item The MLA K cache remains stable along the channel dimension, indicating that a small number of pattern vectors can still effectively cover its distribution.
    \item Along the sequence dimension, MLA exhibits the same gradual evolution behavior as GQA/MHA, which can be attributed to the RoPE information injected into the cache.
\end{itemize}

These findings indicate that PatternKV can be directly applied to the KV cache that is actually stored in MLA, without requiring any additional architectural adaptations. In future work, we plan to conduct more systematic MLA-based experiments to further validate these observations and to explore broader design choices for pattern-based KV compression in non-standard attention architectures.

\end{document}